\documentclass[lettersize,journal]{IEEEtran}
\usepackage{amsmath,amsfonts}
\usepackage{algorithmic}
\usepackage{algorithm}
\usepackage{array}
\usepackage[caption=false,font=normalsize,labelfont=sf,textfont=sf]{subfig}
\usepackage{textcomp}
\usepackage{stfloats}
\usepackage{url}
\usepackage{verbatim}
\usepackage{graphicx}
\usepackage{tabularx}
\usepackage{cite}
\usepackage{subcaption}
\usepackage{subfig}
\usepackage{color}
\usepackage{bibentry}
\usepackage{multirow}
\usepackage{booktabs}
\usepackage{float}
\usepackage{epstopdf}
\usepackage[graphicx]{realboxes}
\usepackage{amsthm,amssymb}
\usepackage{mathrsfs}
\usepackage{makecell}
\usepackage[caption=false,font=normalsize,labelfont=sf,textfont=sf]{subfig}

\usepackage{hyperref}

\newtheorem{definition}{Definition}
\newsavebox\CBox
\def\textBF#1{\sbox\CBox{#1}\resizebox{\wd\CBox}{\ht\CBox}{\textbf{#1}}}
\usepackage{subcaption}

\begin{document}

\title{Mastering the Minority: An Uncertainty-guided Multi-Expert Framework for Challenging-tailed Sequence Learning}

\author{Ye Wang, Zixuan Wu, Lifeng Shen, Jiang Xie, Xiaoling Wang, Hong Yu, and Guoyin Wang
\thanks{This work was sponsored by the National Natural Science Foundation of China (62306056,62136002,62221005).}
\thanks{Ye Wang, Zixuan Wu, Lifeng Shen, Jiang Xie, and Hong Yu are with Key Laboratory of Cyberspace Big Data Intelligent Security, Ministry of Education; School of Artificial Intelligence, Chongqing University of Post and Telecommunications, Chongqing 400065, China.}
\thanks{Xiaoling Wang is with the School of Computer Science and Technology, East China Normal University, Shanghai, 200062, China.}
\thanks{Guoyin Wang is with the Chongqing Key Laboratory of Brain-Inspired Cognitive Computing and Educational Rehabilitation for Children with Special Needs, Chongqing Normal University, Chongqing, 401331, China.}
\thanks{Corresponding author: Hong Yu (yuhong@cqupt.edu.cn) and Lifeng Shen (shenlf@cqupt.edu.cn).}
}

% The paper headers
\markboth{ IEEE TRANSACTIONS ON AUDIO, SPEECH AND LANGUAGE PROCESSING.}%
{Shell \MakeLowercase{\textit{et al.}}: A Sample Article Using IEEEtran.cls for IEEE Journals}

% \IEEEpubid{0000--0000/00\$00.00~\copyright~2021 IEEE}
% Remember, if you use this you must call \IEEEpubidadjcol in the second
% column for its text to clear the IEEEpubid mark.

\maketitle

\begin{abstract}
Imbalanced data distribution remains a critical challenge in sequential learning, leading models to easily recognize frequent categories while failing to detect minority classes adequately. The Mixture-of-Experts model offers a scalable solution, yet its application is often hindered by parameter inefficiency, poor expert specialization, and difficulty in resolving prediction conflicts. To Master the Minority classes effectively, we propose the Uncertainty-based Multi-Expert fusion network (UME) framework. UME is designed with three core innovations: First, we employ Ensemble LoRA for parameter-efficient modeling, significantly reducing the trainable parameter count. Second, we introduce Sequential Specialization guided by Dempster-Shafer Theory (DST), which ensures effective specialization on the challenging-tailed classes. Finally, an Uncertainty-Guided Fusion mechanism uses DST's certainty measures to dynamically weigh expert opinions, resolving conflicts by prioritizing the most confident expert for reliable final predictions. Extensive experiments across four public hierarchical text classification datasets demonstrate that UME achieves state-of-the-art performance. We achieve a performance gain of up to 17.97\% over the best baseline on individual categories, while reducing trainable parameters by up to 10.32\%. The findings highlight that uncertainty-guided expert coordination is a principled strategy for addressing challenging-tailed sequence learning. Our code is available at \url{https://github.com/CQUPTWZX/Multi-experts}.

% Hierarchical text classification (HTC) usually exhibits a long-tailed distribution, which is significantly challenging for the tailed samples, due to the scarcity of samples at lower hierarchy levels. To alleviate the imbalanced issue in HTC, we introduce an Uncertainty-based Multi-Expert fusion network (UME). Specifically, to fully integrate the inherent uncertainty in the multiple-expert decision-making process, the proposed UME employs the Dempster-Shafer evidence theory (DST) to enable automated multi-expert fusion learning. Consequently, initial experts can focus on higher-level tasks, while subsequently added experts can specialize in handling more difficult ones. Furthermore, the proposed UME is computationally efficient due to the use of low-rank structures for experts. Compared to recent competitive baselines HiTIN (ACL'23) and HiAdv (COLING'24), the proposed method reduces the computational cost of trainable parameter numbers by 10.32\% and 7.19\%, respectively. Compared to state-of-the-art baselines, UME achieves an average accuracy improvement of 3.78\% for each challenging-tailed class. Regarding classification performance, our proposed method achieves state-of-the-art results on multiple public real-world HTC datasets. Our code is available at \url{https://github.com/CQUPTWZX/Multi-experts}.
\end{abstract}

\begin{IEEEkeywords}
Sequence learning, Uncertainty-fusion, Multi-expert Learning, Hierarchical Text Classification.
\end{IEEEkeywords}

\section{Introduction}
\IEEEPARstart{I}{n} sequence learning, imbalanced data distributions pose a major challenge, leading to situations where some categories are easily recognized while others are difficult to detect \cite{2023long-tailed_survey}. Consequently, models often achieve high accuracy on frequent categories but perform poorly on rare or minority ones. Overlooking these minority categories can have severe consequences. For example, in medical applications, rare diseases are typically underrepresented in data, and failing to detect them undermines the reliability of evidence-based diagnosis \cite{2024uncommon_diseases}. Similarly, in financial security, anomalous transactions and novel attack patterns occur infrequently, and neglecting their detection can expose systems to significant risks \cite{2022anomaly}. 

\begin{figure}[htp]
\centering 
\subfloat[A three-level label tree is used in the news recommendation scene.]
{
    \centering
    \includegraphics[width=6.5cm]{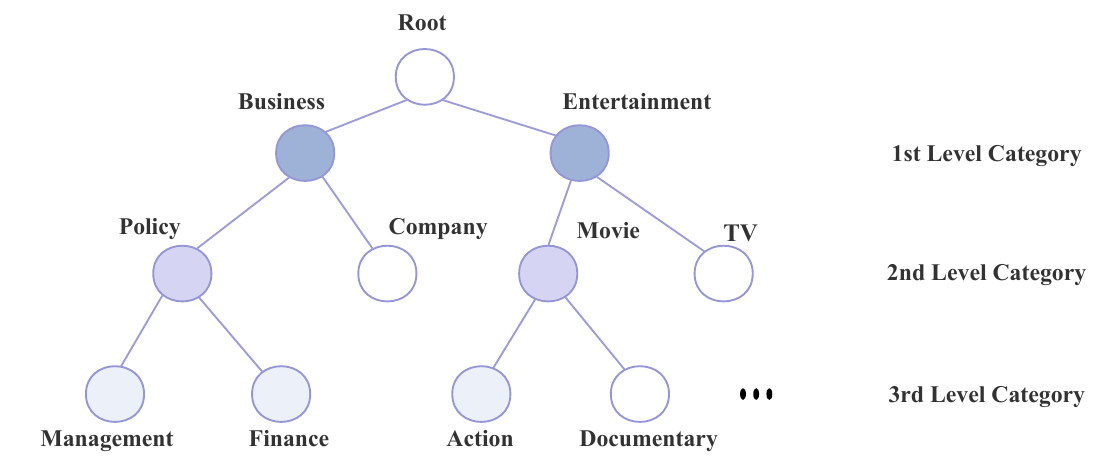}
}
\hfil
\subfloat[Long-tailed label distribution and challenging-tailed performance.]
{
    \centering
    \includegraphics[width=6.5cm]{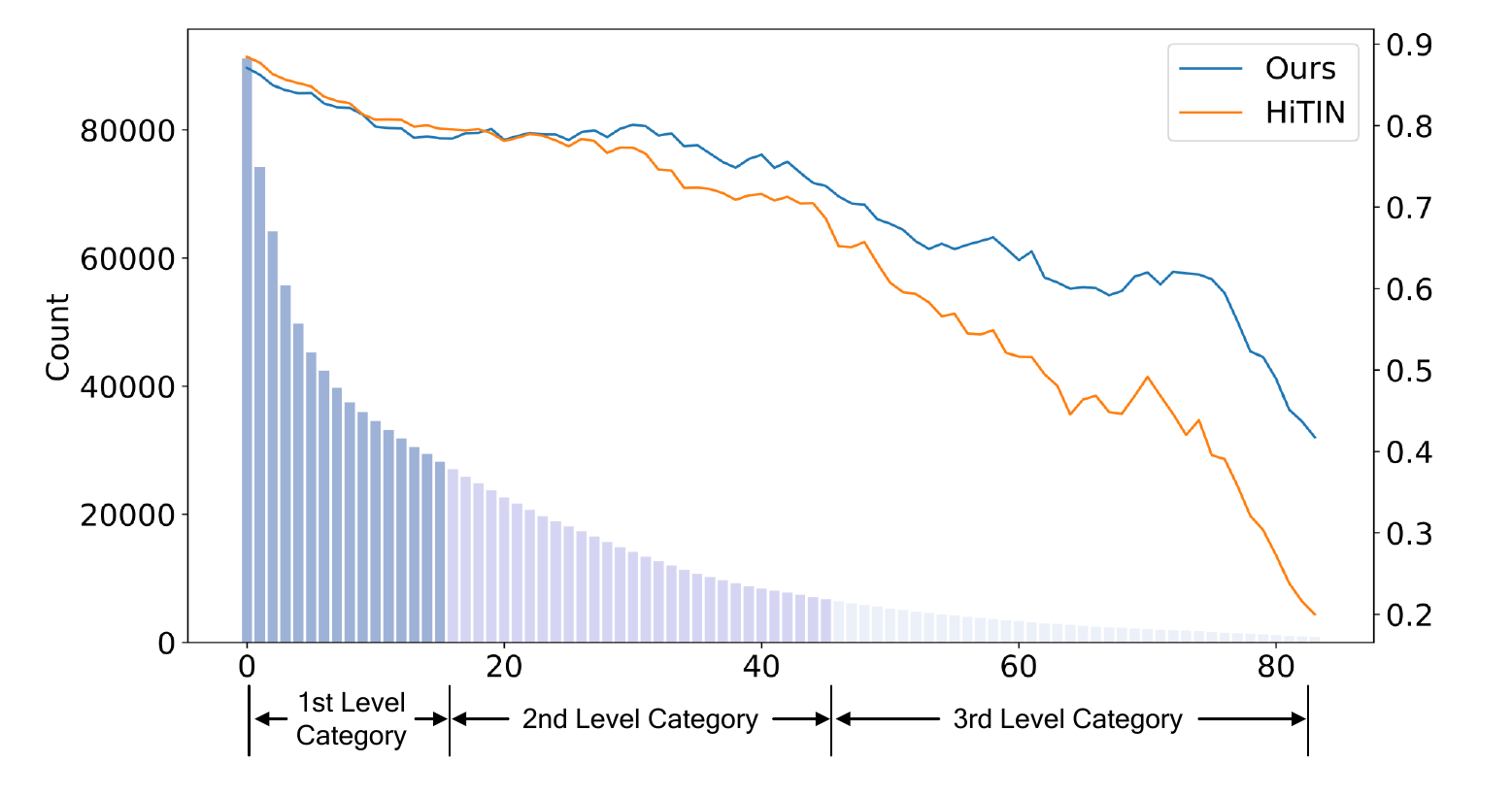}
}
\caption{As the classification hierarchy becomes deeper, the number of samples per label decreases significantly, leading to a pronounced challenging-tailed problem and a sharp drop in the classification performance of minority classes.}
\label{fig:intro}
\end{figure}

The problem caused by imbalanced data distributions becomes particularly pronounced in hierarchical classification tasks, where the data distribution is inherently long-tailed \cite{2021HiMatch,2020Tohre}. Specifically, in Figure \ref{fig:intro}(a), a label tree of three levels is shown from the news recommendation scene, while in Figure \ref{fig:intro}(b), we show the corresponding distribution of hierarchical labels. The objective of sequence learning here is to predict the specific leaf nodes to which a given text belongs. Figure \ref{fig:intro} clearly exhibits that the more specific categories are typically concentrated at the lower levels of the hierarchy, making them harder to learn due to limited training samples. If these categories are overlooked, many leaf nodes in the classification tree will not be correctly covered, which compromises the integrity of the hierarchy and causes the loss of critical fine-grained information \cite{2020Deep-RTC,2021HTTN,2023PA-TRP}.

The Mixture of Experts (MoE) framework has recently emerged as a key technique for addressing data imbalance \cite{2023K-HTC,2020X-Transformer}. It leverages multiple specialized sub-models, each designed to capture distinct patterns or handle specific data subsets. However, the main challenge is to ensure the proper coordination among these experts, which encourages sufficient specialization while maintaining global consistency. Determining how to assign data to the most suitable experts and how to aggregate their outputs remains a critical problem. 

We propose that an effective multi-expert system must address three fundamental challenges. First, parameter inefficiency arises when combining several large experts, leading to a sharp increase in parameters and high training and memory costs. Second, expert specialization is difficult to achieve. Without an explicit mechanism for task allocation, experts may converge to similar patterns, causing redundancy, overfitting, and poor coverage of data diversity. Third, expert opinion conflict appears when experts give inconsistent predictions, requiring a reliable strategy to handle these differences and produce stable results \cite{conflict}. 

To address these challenges, we propose the Uncertainty-based Multi-Expert fusion network (UME). First, we tackle parameter inefficiency with Ensemble LoRA, which leverages Low-Rank Adaptation (LoRA) to enable lightweight multi-expert modeling. Each expert is fine-tuned with only a small number of trainable parameters. Second, to promote expert specialization, we propose a Sequential Specialization strategy based on the Dempster-Shafer evidence theory (DST) \cite{2022DST}. In this strategy, experts are trained one by one rather than at the same time. This approach encourages a natural division of labor. Later experts are specifically trained to focus on samples that earlier experts fail to classify correctly. As a result, experts develop distinct uncertainty profiles: early-stage experts show high uncertainty on difficult samples, while late-stage experts exhibit low uncertainty on the same samples they have been trained to master. Finally, UME integrates these signals through an Uncertainty-Guided Fusion mechanism, which dynamically assigns decision weights based on expert confidence and resolves conflicts by prioritizing the most certain expert for each sample. 

The main contributions of this work are summarized as follows:
\begin{itemize}
    \item[(1)] We propose an efficient and effective multi-expert fusion network named UME, which improves the recognition of challenging-tailed categories in sequential learning.
    \item[(2)] We introduce a sequential specialization mechanism to promote expert diversity. Later added experts focus more on challenging-tailed samples, identifying through their predictive uncertainty.
    \item[(3)] To alleviate conflicts among experts, we leverage DST theory to dynamically integrate multiple experts based on their uncertainty, ensuring consistent and reliable decision fusion.
    \item[(4)] To enhance parameter efficiency, we incorporate low-rank experts, significantly reducing the number of trainable parameters by 10.32\% and 7.19\% compared with recent competitive baselines HiTIN (ACL’23) and HiAdv (COLING’24). Extensive experiments on four public datasets demonstrate that UME achieves state-of-the-art performance, particularly on challenging-tailed categories.
\end{itemize}

\section{Related Work}
\subsection{Hierarchical Text Classification}
Hierarchical Text Classification (HTC) is an important research topic in natural language processing, where collections of documents contain hierarchically structured concepts. Currently, HTC methods evolve into two main approaches, as defined by Silla and Freitas \cite{2011A_survey}. The first is local methods, which train specialized classifiers for specific nodes or levels without considering the entire hierarchy \cite{2019local}. The second is global methods, which utilize the full hierarchical structure through penalties or hierarchy-aware models \cite{2020HiAGM}. Early research mainly focuses on capturing local information, while recent studies emphasize leveraging global information by transferring knowledge across the hierarchy and incorporating it into predictions \cite{2024revisiting}. Additionally, with the advancement of pretrained language models, recent work continues to enhance the overall classification accuracy of HTC models. For example, Zhang et al. \cite{2024HALB} enhance hierarchical text classification performance through multi-label negative supervision and asymmetric loss by constructing a Hierarchy-Aware and Label-Balanced model (HALB). To address the challenges of long-document classification, Liu et al. \cite{2024longMulti} utilized an interactive graph network to align local and discourse-level textual features with label structure information. In contrast, Zhu et al. \cite{2024HILL} present a strategy called Hierarchy-aware Information Lossless contrastive Learning (HILL) that improves HTC performance by retaining both semantic and syntactic information. Even though previous techniques have achieved good performance in HTC, training classification models for tailed labels remains more challenging than for head labels due to the long-tail label distribution, leading to underfitting of tailed labels. Xu et al. \cite{2023long-tail-AAAI} propose the Label-Specific Feature Augmentation (LSFA) framework for text classification to improve tail label representations. It enhances rare labels by generating positive feature-label pairs and transferring intra-class semantic variations from head labels using a prototype-based VAE. Furthermore, challenging the reliance on a fixed label order, Yan et al. \cite{2023order} explored a random generative method to learn the label hierarchy, demonstrating that a predefined order may not always be optimal.

However, their approach does not account for the impact of hierarchy in addressing the underfitting issue of tailed labels. The hierarchy leverages multi-level semantic aggregation to build semantic bridges, enabling tail labels to learn richer context from head labels and enhance their representations. Thus, we propose to leverage multi-expert ensembles to improve classification accuracy by allowing experts to specialize in different tasks. By incorporating uncertainty-based weighting among experts, our method provides better support for challenging tail-label samples at lower levels of the hierarchy.

\subsection{Uncertainty-based Text Classification} 
%In recent years, uncertainty-based methods have been increasingly explored to enhance text classification performance. For example, Mukherjee et al. \cite{mukherjee2020uncertainty} improve self-training for few-shot text classification by leveraging Bayesian deep learning to estimate uncertainty and select reliable data. Similarly, Yao et al. \cite{yao2019clinical} incorporate rule-based features and knowledge-guided convolutional neural networks to enhance text classification. Building on this, Li et al. \cite{2022trustworthy} propose an uncertainty estimation method to identify hard samples in text classification. To address out-of-distribution issues, Hu et al. \cite{hu2021uncertainty} introduce an uncertainty-aware reliable network that applies evidential uncertainty for out-of-distribution detection. Although these approaches effectively utilize uncertainty in text classification, they cannot be directly applied to HTC. 

%Unlike standard text classification, HTC requires a more precise modeling of the label hierarchy structure to capture complex dependencies between labels. As the hierarchy deepens, both classification difficulty and uncertainty increase. To address this, we introduce uncertainty within the evidence theory framework. By measuring this uncertainty, we enhance both the reliability and efficiency of the classification process.

In recent years, uncertainty-aware methods have been increasingly explored to enhance text classification performance. For example, Mukherjee et al. \cite{mukherjee2020uncertainty} leverage Bayesian deep learning to estimate uncertainty for few-shot text classification. Li et al. \cite{2022trustworthy} propose uncertainty-based criteria to identify hard samples during training, while Hu et al. \cite{hu2021uncertainty} introduce an evidential uncertainty network for out-of-distribution detection. These studies demonstrate the effectiveness of Bayesian or evidential uncertainty modeling in improving robustness.

While Bayesian approaches provide a principled treatment of uncertainty, many sampling-based methods (e.g., Monte Carlo Dropout \cite{gal2016dropout}) require repeated stochastic forward passes at inference time, leading to increased computational cost. In contrast, we employ Dempster–Shafer Theory to model uncertainty directly from output evidence. Crucially, DST explicitly represents conflict between experts, allowing us to distinguish ignorance from disagreement. This capability is essential for our sequential fusion strategy, yet remains underexplored in long-tailed hierarchical text classification.

\subsection{Multi-Expert Text Classification}
Multi-expert learning fuses multiple single experts' classification results to improve the final performance \cite{2023K-HTC,2021SLR}. For example, Zhao et al. \cite{2021hierarchical-long-tail} propose a multi-expert method that assigns different models to learn global and local features separately, allowing for better classification of long-tailed data using a hierarchical approach. Subsequently, the optimal decision threshold within the multi-expert method \cite{2023rebalanced} is proposed, introducing a gating network to enhance the performance in classification. 

The Divide, Conquer, and Combine paradigm has been explored in sequence learning. For instance, Wang et al. \cite{wang2023divide} utilize it for zero-shot dialogue tracking. They explicitly partition data into semantic clusters and combine experts via distance-based mapping. In contrast, our UME framework adapts this paradigm for long-tailed classification with distinct mechanisms. Instead of semantic clustering, we employ Sequential Specialization to divide the problem by sample difficulty. Furthermore, rather than distance-based inference, we utilize Dempster-Shafer Theory to combine experts. This approach specifically addresses the issue of conflicting expert predictions in imbalanced scenarios.

However, regarding multi-expert in HTC, the issue occurs when different experts have varying opinions on the classification results, causing conflicts in the overall multi-expert ensemble outcomes. Compared to previous approaches, the proposed method provides a more flexible solution for HTC, particularly in addressing the challenge of long-tailed category classification. In multi-expert networks, experts may produce conflicting opinions, making it crucial to effectively integrate their predictions. To tackle this, we propose to introduce an uncertainty-based multi-expert fusion framework that leverages DST for dynamic expert weighting.

\begin{figure*}[t!]
\centering 
\includegraphics[width=17cm]{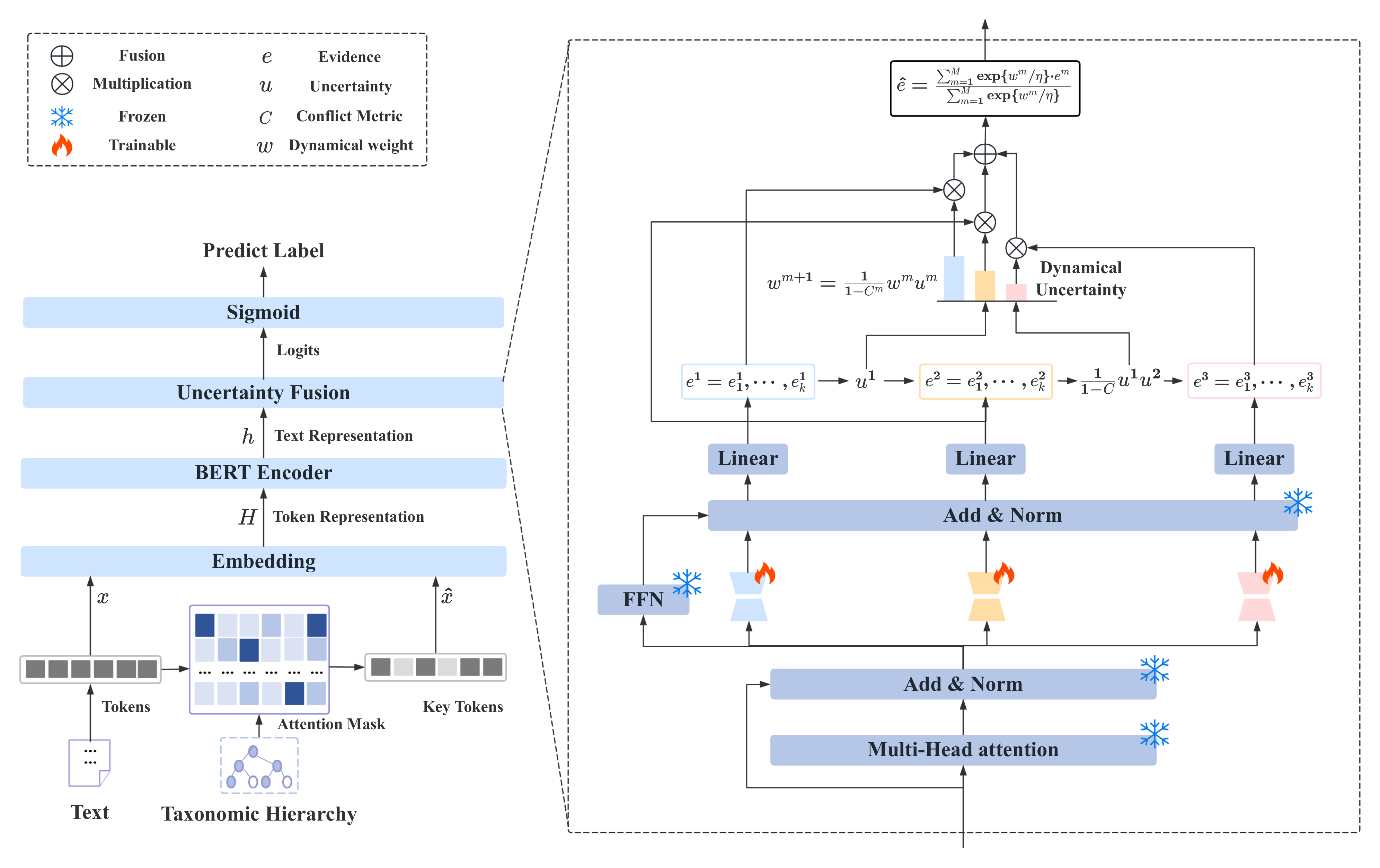}
\caption{The structure of our proposed model UME. The uncertainty of every single expert is attained for multi-expert learning. Multi-expert Joint Uncertainty is dynamically ensembled by the degree of uncertainty among experts.}
\label{fig:model}
\end{figure*}

\section{Methodology}
In this Section, we formally elaborate on the proposed Uncertainty-based Multi-Expert fusion network (UME). We first introduce the Task Definition and the formal definitions of Evidence and Uncertainty in Section 3.1 Preparation. Section 3.2 introduces how to obtain representations of Text and Label Hierarchy, while proposing the Low-rank Experts constructed in Section 3.3. In our multi-expert framework, each expert extracts evidence from the input and formulates a classification opinion. We then construct the Uncertainty-based Multi-Expert Fusion Network (UME) in Section 3.4. The main workflow is illustrated in Figure \ref{fig:model}, with the fusion process comprising three key steps (detailed in Section 3.4): i) uncertainty quantification; ii) uncertainty combination; and iii) likelihood-based optimization. Section 3.5 presents our training objective, which demonstrates how experts dynamically participate during training.

\subsection{Preliminaries}
In the HTC task, we are given a label set $L = \{ l_1^1,l_2^1, \cdots,l_K^H\}$. organized into a hierarchical structure, typically represented as a label tree. Here, $H$ is the height of the hierarchical label tree, and $K$ denotes the total number of labels in the set. The goal of HTC is to predict a subset of this label set for a given input text $S = \{ {c_1},{c_2}, \cdots,{c_N}\},$ where $c_i(i=1,\dots,N)$ represents the input tokens, and $N$ is the total number of tokens in the text. The task aims to assign the appropriate labels to the input text based on its content while considering the hierarchical relationships between labels. In the following, we provide the formal definitions of Evidence and Uncertainty used in the proposed fusion framework.

\begin{definition}\label{def:evidence}\textbf{Evidence.} In neural networks, logits are the outputs of the final layer in classification tasks and represent the model's raw predictions before applying functions like softmax. These logits reflect the model's confidence in each class. In Dempster-Shafer Theory (DST), evidence refers to the degree of support for a specific proposition, such as a sample belonging to a particular category. We treat logits as evidence because they provide a quantifiable measure of support for each class. Formally, we define evidence $e=logits$ where $e = [{e_1},{e_2}, \cdots ,{e_K}]$ for a total of K classes. This mapping aligns well with DST's requirement for quantifiable support, as the logits directly represent the model's assessment of how likely a sample belongs to each category. By using logits as evidence, we bridge the outputs of neural networks with DST, enabling the integration of uncertainty modeling and evidence fusion into our framework. This allows the model to better handle challenging classification tasks, such as those involving long-tailed distributions. 
\end{definition}

\begin{definition}\label{def:uncertainty}\textbf{Uncertainty.} Subjective Logic (SL) formalizes DST’s notion of belief assignments over a frame of discernment as a Dirichlet Distribution \cite{2018uncertainty}. Hence, it allows one to use the principles of evidential theory to quantify belief masses and uncertainty through a well-defined theoretical framework. SL considers a frame by assigning belief mass ${b_k}$ to the k-th class, $k=1,\dots,K$, thereby determining the overall uncertainty mass $u$. These $K+1$ mass values are all non-negative and sum to 1, ensuring a complete and consistent distribution of belief and uncertainty. Uncertainty quantifies the trustworthiness of the classification results by distributing belief mass ${b_k} = \frac{{{\alpha _k} - 1}}{S}$ across possible labels. Here, ${\alpha _k}$ represents the Dirichlet parameters for the k-th class, and $S = \sum\nolimits_{k = 1}^K {{({e_k} + 1)}}$ is the Dirichlet strength, reflecting the total amount of evidence. A larger $S$ indicates more accumulated evidence, reducing uncertainty and leading to more confident belief assignments. When no evidence is present (${e_k} = 0$ for all $k$), the belief for each class is zero, and uncertainty reaches its maximum (i.e., $u = 1$). The Dirichlet parameters are calculated as ${\alpha _k} = {e_k} + 1$, where ${e_k}$ is the evidence for the k-th class, adding 1 prevents zero belief when no evidence is observed and ensures a smooth uncertainty estimation. Evidence ${e_k}$ quantifies the model's support for each class. A higher evidence value  ${e_k}$ results in a larger Dirichlet parameter ${\alpha _k}$, leading to increased belief mass ${b_k}$, which corresponds to greater confidence in the class prediction. Conversely, lower evidence implies greater uncertainty. The unallocated belief mass $u$, which represents uncertainty, is defined as:
\begin{equation}\label{u}
u=1-\sum\nolimits_{k=1}^{K}{{{b}_{k}}}=\frac{K}{S},
\end{equation}
where $S = \sum\nolimits_{k = 1}^K {{\alpha _k}}$. This ensures that:
\begin{equation}\label{1}
u + \sum\nolimits_{k = 1}^K {{b_k}}  = 1,
\end{equation}
where $u$ captures the system's lack of confidence in all categories. In summary, higher evidence leads to higher belief mass ${b_k}$ and lower uncertainty $u$, whereas lower evidence results in increased uncertainty.
\end{definition}

\subsection{Representations of Text and Label Hierarchy}
Text representation denoted by $H$ is a hidden representation of the input text, while key tokens' representation $\hat{H}$ refer to the representation related to the hierarchical structure, obtained by combining it with the graph encoder. We first utilize the Embedding layer of BERT \cite{kenton2019bert}:
\begin{equation}\label{eq1}
H = {\rm{Embedding}}(x),
\end{equation}
where $H \in {R^{n \times {d_h}}}$ represents hidden representation and ${d_h}$ describe the dimension of the hidden layer; $x$ defines input token sequence. Meanwhile, \cite{2021graphormer} is used as the graph encoder with the guidance of labels. Here, key tokens are constructed by masking unimportant tokens. Combining and contrasting the two representations can inject hierarchical information into the BERT Encoder. Following the approach of HGCLR \cite{2022contrast}, we adopt Gumbel-Softmax \cite{2017gumbel} instead of the standard Softmax function to make the sampling operation differentiable. Tokens for which the probability of belonging to label ${y_i}$ exceeds the threshold $\gamma$ are constructed as key tokens $\hat x$:
\begin{equation}\label{eq11}
\hat x = \left\{ {{x_i},{\rm{ \quad if \quad }}\sum\limits_{j \in y} {{P_{ij}}} > \gamma; {\rm{ \quad else \quad  0}}} \right\},
\end{equation}
where ${P_{ij}}$ represents the probability of sampling. Similarly, the key tokens undergo the same embedding layer to obtain its corresponding representation $\hat{H} = {\rm{Embedding}}(\hat{x})$.

\subsection{Low-rank Experts}
For BERT Encoder, both $H$ and $\hat{H}$ are taken as inputs for calculating the token representations and hierarchy representations. Note that $\hat{H}$ is used to enhance $H$'s representation via contrastive learning that will be introduced in Equation (\ref{sec:cl}). We freeze the parameters of the Transformer blocks in the BERT Encoder and subsequently apply uncertainty fusion using the LoRA technique in the last FFN layer of the BERT Encoder. Specifically, we fine-tune the FFN layer using $M$ low-rank experts defined by:
\begin{equation}\label{eq:LoRA}h_{out} = {W_0}h + \Delta Wh = {W_0}h + \sum\limits_{m = 1}^M {{B_m}{A_m}h},\end{equation}
where $h$ denotes the input hidden state to the FFN layer. ${W_0}$ represents the original parameter matrix. $\Delta W$ is the matrix that needs to be updated in the original FFN layer. $A_m$ and $B_m$ denote the low-rank decomposition matrices of $W_0$ for the $m$-th expert. $A_m$ is initialized using random Gaussian values, while $B_m$ is initialized to zero.

\subsection{Uncertainty-based Multi-Expert Fusion Network}
Given an input sample, each expert produces evidence scores for hierarchical labels. These scores are converted into uncertainty-aware representations using Dempster–Shafer Theory. We then compute prediction conflict and uncertainty between adjacent experts, which are used to dynamically determine each expert’s contribution. Experts with high confidence dominate simple samples, while high-conflict samples are progressively routed to later experts for further refinement. Finally, weighted evidence from all experts is fused to produce the final prediction.

\textbf{(i) Uncertainty quantification} serves as the foundation for integrating multiple experts. We employ Dempster-Shafer Theory (DST) \cite{2018uncertainty} to model hierarchical label prediction, distinguishing between belief mass $b_k$ (evidence) and uncertainty $u$ \cite{2018evidential}. Specifically, we model the belief mass using a Dirichlet distribution $D\left( \cdot \right)$ \cite{2004distribution}:
\begin{equation}\label{eq2}
\begin{aligned}
D(p|\alpha ) = \left\{ \begin{array}{l}\frac{1}{{B(\alpha )}}\prod\nolimits_{k = 1}^K {p_k^{{\alpha k} - 1}} {\rm{ \quad  for \;}}p \in {S_K}\\{\rm{0 \qquad \qquad \quad \quad \quad ; otherwise}}\end{array} \right.,
\end{aligned}
\end{equation}
where $\alpha$ represents the distribution parameters, $B\left( \cdot \right)$ is the beta function, and ${S_K} = \left\{ {p|\sum\nolimits{k = 1}^K {{p_k} = 1\enspace{\rm{ ,}}\enspace0 \le {p_k} \le 1} } \right\}$ denotes the $K$-dimensional unit simplex. Based on this distribution, evidence-based uncertainty is quantified according to Definitions \ref{def:evidence} and \ref{def:uncertainty}.

\textbf{(ii) Uncertainty fusion} combines the uncertainty $u=\{u^1, \dots, u^M\}$ and evidence $e=\{e^1, \dots, e^M\}$ from all experts to dynamically weight their contributions. This mechanism adjusts expert influence based on sample complexity: prioritizing preceding experts for simple samples while leveraging the full ensemble for hard, long-tail categories. First, we sequentially fuse expert uncertainties $u$:
\begin{equation}\label{eq3}
u = {u^1} \oplus {u^2} \oplus  \cdots  \oplus {u^M} = \frac{{\prod\nolimits_{m = 1}^M {{u^m}} }}{{\prod\nolimits_{m = 1}^M {\left( {1 - {C^m}} \right)} }},
\end{equation}
where ${C^m} = \sum\nolimits_{i \ne j} {b_i^m} b_j^{m - 1}$ represents the conflict metric between adjacent experts (with ${C^{\rm{1}}} = {\rm{0}}$), and ${b_i^m}$ denotes the confidence of the $m$-th expert for the $i$-th category. A conflict metric of 0 indicates consistency. To promote diversity, we employ a \textit{Sequential Specialization} strategy where subsequent experts explicitly focus on hard samples characterized by high predictive uncertainty.
Based on the conflict $C^m$ and uncertainty $u^m$, we define the dynamical weight for each expert:
\begin{equation}\label{eq:expert_weight}
{w^{m + 1}} = {w^m} \oplus {u^m} = \frac{1}{{1 - {C^m}}}{w^m}{u^m}.
\end{equation}
Here, ${w^m}$ quantifies the accumulated uncertainty passed from previous experts to expert $m$ (initialized as ${w^1} = 1$, ${w^2} = {u^1}$). These dynamical weights are then applied to aggregate the evidence from all $M$ experts:
\begin{equation}\label{eq5}
\hat{e} = \frac{{\sum\nolimits_{m = 1}^M {{\rm{exp\{ }}{w^m}{\rm{/}}\eta {\rm{\} }} \cdot {e^m}} }}{{\sum\nolimits_{m = 1}^M {{\rm{exp\{ }}{w^m}{\rm{/}}\eta {\rm{\} }}} }},
\end{equation}
where $\eta$ is a temperature factor used to scale the weights via a Softmax-like operation. During inference, the combined evidence $\hat{e}$ is converted into probabilities using the Sigmoid function:
\begin{equation}\label{sigma}
\sigma (\hat{e}) = \frac{1}{{1 + \exp ( - \hat{e})}}.
\end{equation}
Finally, labels with a probability $\sigma (\hat{e})$ exceeding a predefined threshold are selected as predictions. The following section details how these dynamical weights guide the training process.

\textbf{(iii) Likelihood-based Optimization.} We propose to improve the uncertainties from each single expert and combined multi-expert via likelihood-based optimization within the multi-expert system. Before getting the evidence ${{\bf{e}}_{\bf{i}}}$ and the one-hot class vector ${{\bf{y}}_{\bf{i}}}$, the adjusted Dirichlet distribution $\widetilde D({{\bf{p}}_{\bf{i}}}|{{\bf{e}}_{\bf{i}}})$ is used as a priori of the multiple likelihood $P({{\bf{y}}_{\bf{i}}}|{{\bf{p}}_{\bf{i}}})$. Then the negative logarithm of the marginal likelihood is given by
\begin{equation}\label{eq6}
\mathcal{L}_{ml} =  - \log \left[ {\int {\prod\limits_{k = 1}^K {p_{ik}^{{y_{ik}}}} \frac{1}{{B\left( {{e_i}} \right)}}} \prod\limits_{k = 1}^K {p_{ik}^{{e_{ik}} - 1}} d{p_i}} \right]
\end{equation}
This means that the correct class can receive more evidence. Additionally, to solve the problem of high overall evidence of wrong labels, we introduce ${\mathcal{{L}}_{evidence}}$ defined as
\begin{equation}\label{eq7}
\mathcal{L}_{evidence} = KL\left( {D\left( {{p_i}|{{\tilde \alpha }_i}} \right)\parallel D\left( {{p_i}|1} \right)} \right), \end{equation} 
where ${\tilde \alpha _i} = {\rm{1 + }}\left( {{\rm{1}} - {y_i}} \right) \odot {e_i}$ indicates the adjusted Dirichlet parameter. By reducing the difference between the target distribution and the adjusted distribution, the evidence for incorrect classes is diminished. Thus, the objective goal of the single expert is given by :
\begin{equation}\label{eq:old_single}
{\mathcal{L}_{{\rm{single}}}} = \mathcal{L}_{ml} + {\lambda}\left( t \right){\mathcal{L}_{evidence}},
\end{equation}
where ${\lambda}(t) = \min\{ 1,t/T\} $ is the annealing factor (t is the current epoch). Intuitively, when $t$ is larger, the effect of KL divergence is fully applied. This ensures that experts can capture differentiated evidence in the early stages. Finally, the joint goal of dynamic learning evidence in the multi-expert network is given:
\begin{equation}\label{eq:single}
\mathcal{L} = \sum\limits_{i = 1}^N {\sum\limits_{m = 1}^M {1\left\{ {w_i^m > \varepsilon } \right\}} } {\mathcal{L}_{{\rm{single}}}}.
\end{equation}

\subsection{Training Objectives}
Apart from the training loss of a single expert in Equation (\ref{eq:single}), we introduce two losses to enhance the optimization of the multi-expert fusion network. They are 1) Classification loss ${\mathcal{L^C}}$; and 2) Contrastive learning loss ${\mathcal{L}^{con}}$ for enhanced text representation.

\noindent \textbf{Classification loss ${\mathcal{L^C}}$.}
The probability of text $x_i$ appearing on label $j$ is:
\begin{equation}\label{eq:sigmoid}
{p_{ij}} = sigmoid{({W_c}{h_i} + {b_c})_j},
\end{equation}
where ${W_C} \in {R^{k \times {d_h}}}$ and ${b_c} \in {R^k}$ illustrate weights and deviations, respectively. $ h = Enconder(H) ,\ \hat{h} = Enconder(\hat{H})$ represents the text representations and hierarchical representations after passing through the Encoder layer. We employ a binary cross-entropy loss function:
\begin{equation}\label{eq13}
\mathcal{L}_{ij}^\mathcal{C} =  - {y_{ij}}\log ({p_{ij}}) - (1 - {y_{ij}})\log (1 - {p_{ij}}),
\end{equation}
\begin{equation}\label{eq14}
{\mathcal{L^C}} = \sum\limits_{i = 1}^N {\sum\limits_{j = 1}^k {\mathcal{L}_{ij}^\mathcal{C}} }, 
\end{equation}
\noindent where ${y_{ij}}$ is the real label; ${\hat {h_i}}$ is used instead of ${h_i}$, and the classification loss of the key sample signifies ${\hat {\mathcal{L^C}}}$.

\noindent\textbf{Contrastive learning loss ${\mathcal{L}^{cl}}$.}
To calculate the contrastive learning loss, we construct negative samples for tokens and key tokens:
\begin{equation}\label{eq15}
{c_i} = {W_2}F({W_1}{h_i})\;,\;{\hat {c_i}} = {W_2}F({W_1}{\hat {h_i}}).
\end{equation}
Here, ${W_1} \in {R^{{d_h} \times {d_h}}}$, ${W_2} \in {R^{{d_h} \times {d_h}}}$, $F$ takes the ReLU function and ${d_h}$ expresses the number of hidden layers. We define $z \in \{ {c_i}\}  \cup \{ {\hat c_i}\} $, and the NT-Xent loss function\cite{2020NT-Xent} is used to force the distance between the positive and negative examples to become larger, and the NT-Xent loss of ${z_i}$ is calculated:
\begin{equation}\label{sec:cl}
\mathcal{L}_i^{cl} =  - \log \frac{{\exp (sim({z_i},{z_j})/\tau )}}{{\sum\nolimits_{k = 1,k \ne i}^{2N} {\exp (sim({z_i},{z_k})/\tau )} }},
\end{equation}
where $sim$ is the cosine similarity function and $\tau $ is the temperature hyperparameter. ${z_i}$ and ${z_j}$ are mutually real and key tokens. Then, the contrast loss is averaged by:
\begin{equation}\label{eq17}
{\mathcal{L}^{cl}} = \frac{1}{{2N}}\sum\limits_{m = 1}^{2N} {\mathcal{L}_i^{cl}}.
\end{equation}
Thus, the objective of a single expert is updated based on the previous equation \ref{eq:old_single} to:
\begin{equation}\label{eq19}
{\mathcal{\tilde{L}}_{{\rm{single}}}} = \mathcal{L}_{ml} + {\lambda _{kl}}\left( t \right){\mathcal{L}_{evidence}} +{\mathcal{L^C}} + {\hat {\mathcal{L^C}}} + {\mathcal{L}^{cl}}.
\end{equation}
Finally, the overall training objective is:
\begin{equation}\label{eq:total loss}
\mathcal{{L}}_{final} = \sum\limits_{i = 1}^N {\sum\limits_{m = 1}^M {1\left\{ {w_i^m > \varepsilon } \right\}} } {\mathcal{\tilde{L}}_{{\rm{single}}}}.
\end{equation}
\section{Experiments}
In this section, we conduct experiments to verify the effectiveness of the proposed multiple-expert fusion network for hierarchical text classification. Specifically, we will answer the following research questions:
\begin{itemize}
    \item[\textbf{RQ1}:] Does the proposed UME outperform the recent strong baselines?
    \item[\textbf{RQ2}:] What is the performance on challenging-tailed samples?
    \item[\textbf{RQ3}:] What are the effects of the number of experts?
    \item[\textbf{RQ4}:] How multiple experts handle challenging-tailed samples?
    \item[\textbf{RQ5}:] What are the effects of components and fusion strategies in UME?
    \item[\textbf{RQ6}:] Is it still efficient by using multiple-expert fusion?
    \item[\textbf{RQ7}:] Can LLMs replace the proposed supervised framework in hierarchical text classification?
\end{itemize}

\subsection{Datasets} 
We select four public HTC datasets in the experiments: Web of Science (WOS) \cite{2017WOS}, RCV1-V2 \cite{2004rcv1}, AAPD \cite{AAPD}, and BGC \footnote{BGC dataset is available at: \url{www.inf.uni-hamburg.de/en/inst/ab/lt/resources/data/blurb-genre-collection.html}}. Specifically, WOS and AAPD contain abstracts of papers published in the Web of Science database and Arxiv respectively, along with the corresponding subject categories. RCV1-V2 is a news classification corpus, while the BGC dataset consists of book introductions and metadata. WOS is used for single-path hierarchical text classification, while RCV1-V2, AAPD, and BGC include multi-path classification labels. Statistical details are listed in Table \ref{tab:data}.

\begin{figure}[t!]
    \centering
    % 第一行：WOS 和 RCV1
    \subfloat[WOS]{
        \includegraphics[width=0.47\linewidth]{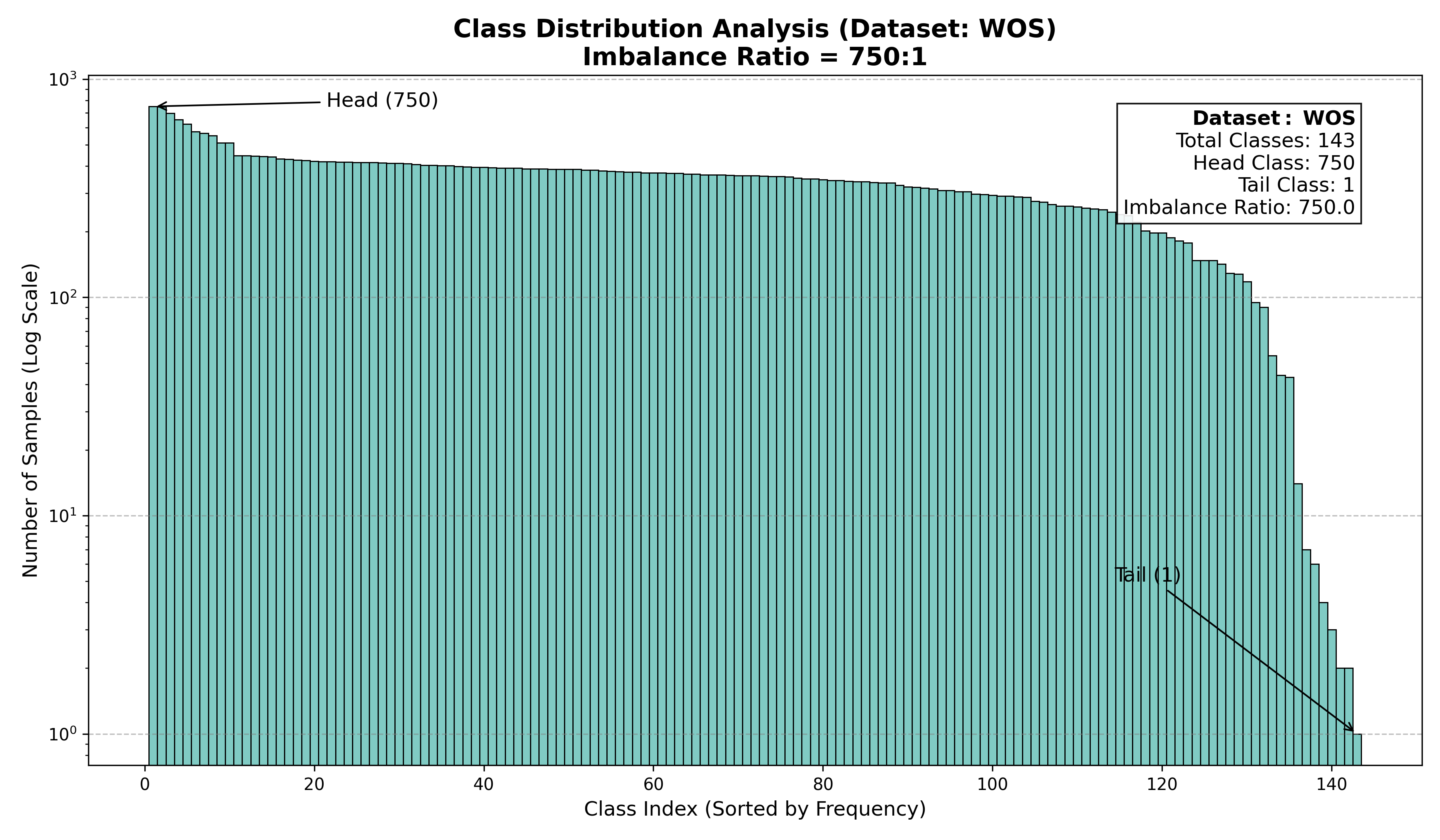}
    }
    \hfil
    \subfloat[RCV1-V2]{
        \includegraphics[width=0.47\linewidth]{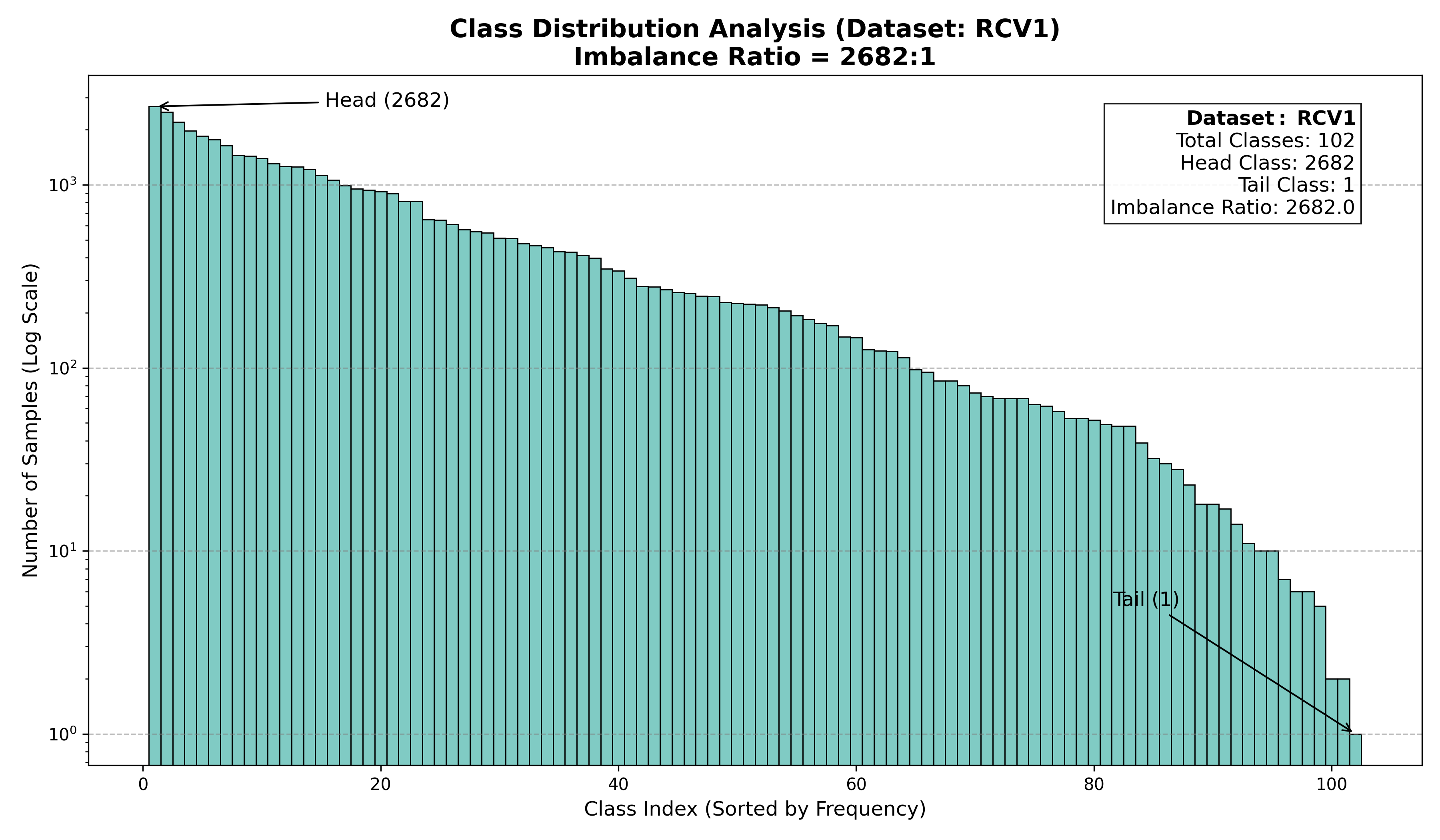}
    }
    \\ % 换行，开始第二行
    % 第二行：AAPD 和 BGC
    \subfloat[AAPD]{
        \includegraphics[width=0.47\linewidth]{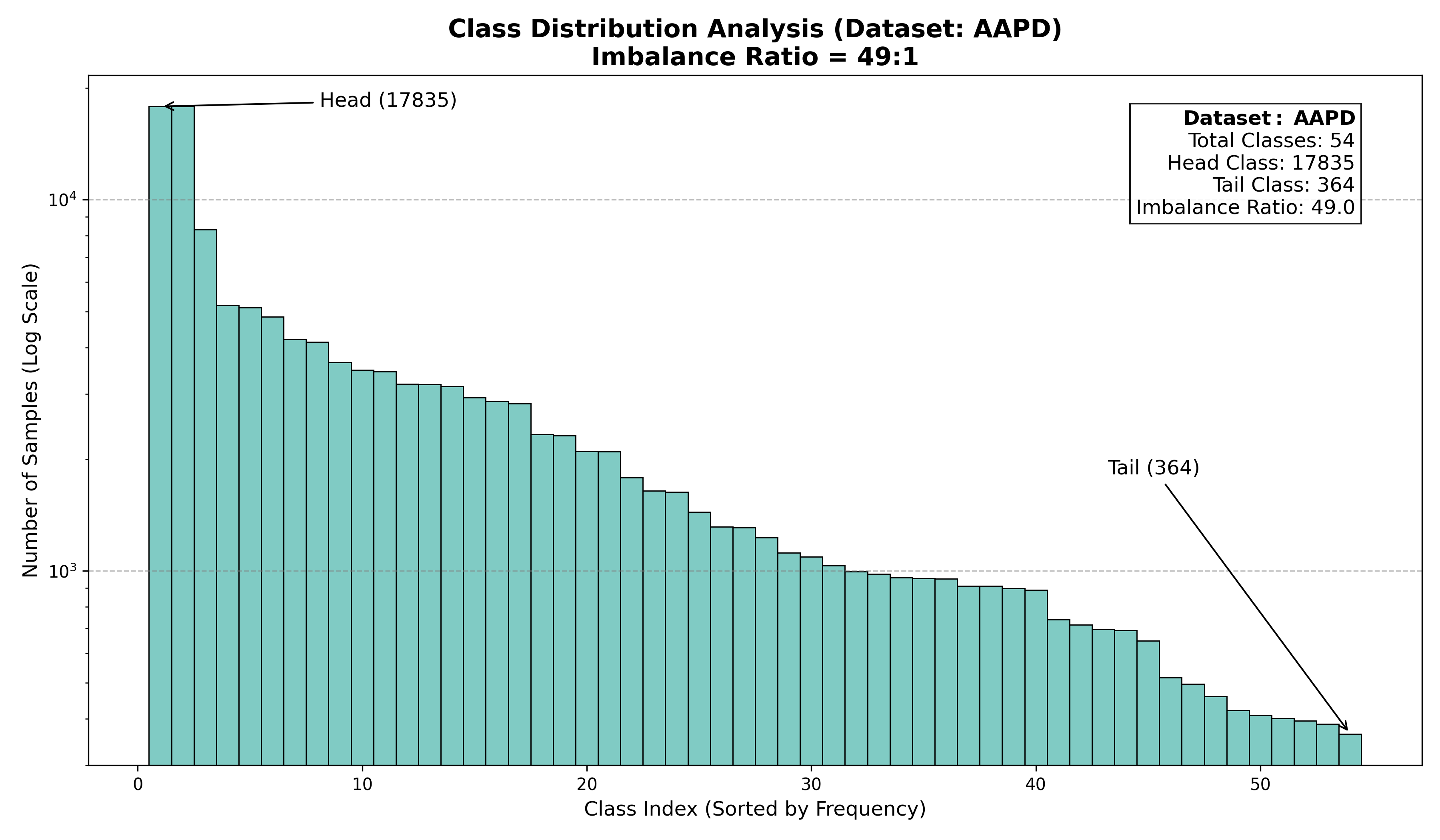}
    }
    \hfil
    \subfloat[BGC]{
        \includegraphics[width=0.47\linewidth]{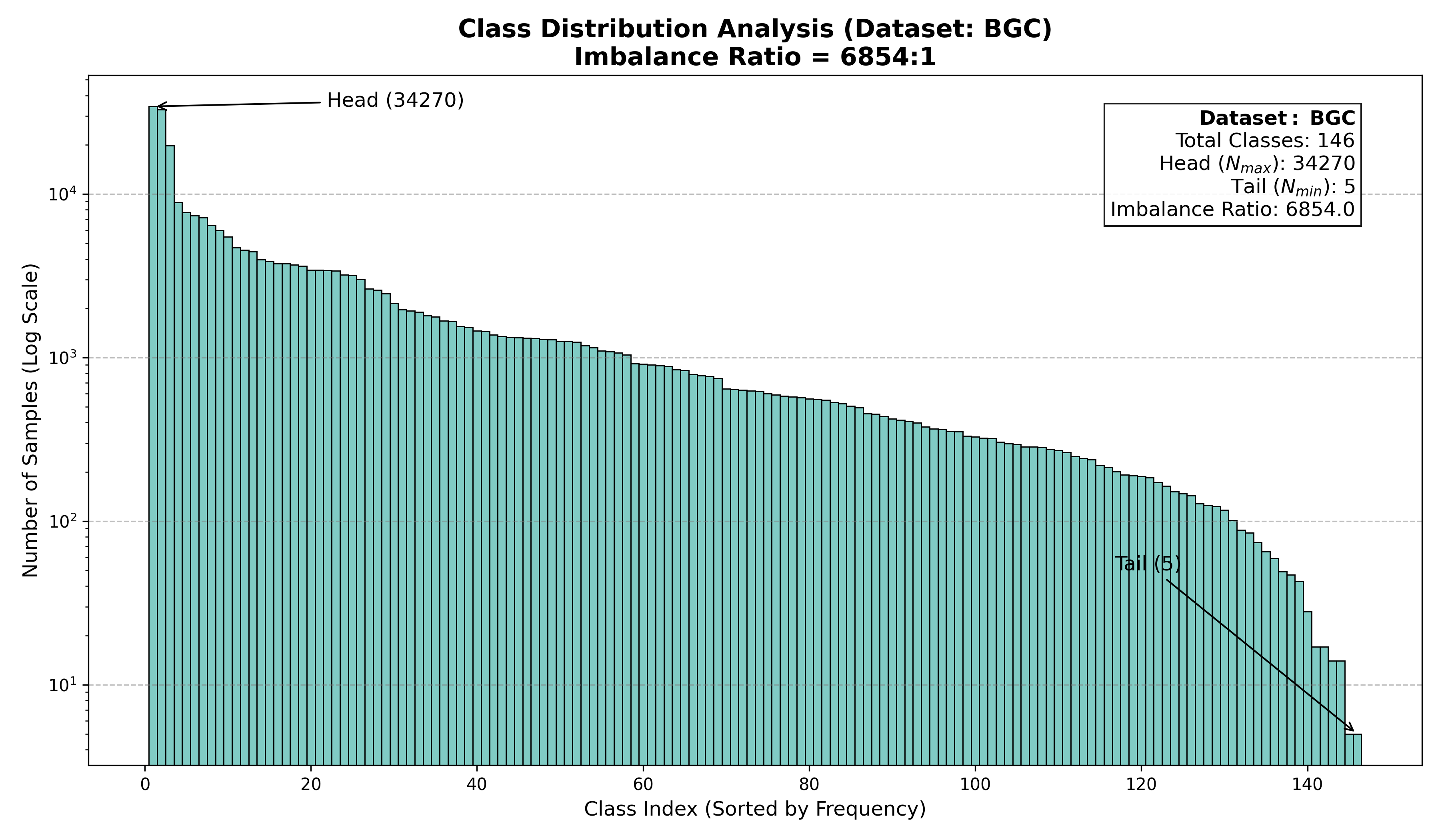}
    }
    % 图例说明
    \caption{Class distribution analysis across four benchmark datasets. The histograms use a logarithmic scale to visualize the severe long-tailed nature and extreme class imbalance ratios (IR) present in WOS, RCV1-V2, AAPD, and BGC.}
    \label{fig:data_distribution}
\end{figure}

We conduct a statistical analysis on the class distributions of four benchmark datasets: AAPD, WOS, RCV1, and BGC. As illustrated in Figure \ref{fig:data_distribution}, all datasets exhibit severe long-tailed characteristics with distinct degrees of imbalance. AAPD shows a moderate imbalance ratio (IR) of 49.0. In contrast, WOS and RCV1 reveal extreme data sparsity in tail classes, where the minimum sample count ($N_{min}$) drops to 1, resulting in high IRs of 750.0 and 2682.0, respectively. BGC presents the most significant disparity, with an IR reaching 6854.0, where the head class contains over 34,000 samples while the tail class has only 5. These statistics quantitatively confirm the intrinsic difficulty and the severe class imbalance issue present in these datasets.

\begin{table}[t]
\caption{The statistical details of datasets. $ \left | Y \right | $ represents the number of unique labels in each dataset. $ Avg\left ( y_{i} \right ) $  the average value of the labels for each entry across the dataset. \textit{Depth} indicates the depth of the hierarchical structure in the dataset. The last three columns of the table include the number of instances in the training, validation, and testing sets, respectively.}
\label{tab:data}
\centering
\begin{tabular}{ccccccc}
\midrule
Dataset & $ \left | Y \right | $  & $ Avg\left ( y_{i}  \right ) $  & Depth & \# Train & \# Dev & \# Test \\ \midrule
WOS     & 141 & 2.0     & 2     & 30,070   & 7,518  & 9,397   \\
RCV1-V2     & 103 & 3.24    & 4     & 18,520   & 4,629  & 781,265 \\
AAPD & 54 & 2.41    & 2     & 43,872   & 10,968  & 1,000 \\
BGC & 146 & 3.01 & 4 & 58,800 & 14,700 & 18,394 \\ \midrule
\end{tabular}
\end{table}

\begin{table*}[t!]
\caption{The comparison of different models on WOS, RCV1-V2, AAPD, BGC and NYT.}
\label{tab:ALL_MODELS}
\centering
\scriptsize
\tabcolsep=0.12cm
\renewcommand\arraystretch{1.2}
\begin{tabular}{lcccccccccc}
\midrule
\multirow{2}{*}{Model} & \multicolumn{2}{c}{WOS} & \multicolumn{2}{c}{RCV1-V2} & \multicolumn{2}{c}{AAPD} & \multicolumn{2}{c}{BGC} & \multicolumn{2}{c}{NYT} \\ \cline{2-11}
\rule{0pt}{8pt}
& Micro-F1 & Macro-F1 & Micro-F1 & Macro-F1 & Micro-F1 & Macro-F1 & Micro-F1 & Macro-F1 & Micro-F1 & Macro-F1 \\ \midrule
\multicolumn{11}{c}{\textbf{Hierarchy-Aware Models}} \\ \hline
HiAGM (ACL2020) & 85.11$\pm$1.4 & 80.02$\pm$0.7 & 83.17$\pm$1.8 & 62.35$\pm$1.9 & 75.54$\pm$2.1 & 58.86$\pm$1.2 & 76.61$\pm$0.9 & 57.27$\pm$0.7 & \textemdash & \textemdash \\
HTCInfoMax (NAACL2021) & 84.86$\pm$0.6 & 79.12$\pm$0.2 & 83.26$\pm$0.8 & 60.66$\pm$1.5 & 77.42$\pm$0.7 & 57.72$\pm$0.7 & 74.26$\pm$0.4 & 56.81$\pm$1.0 & \textemdash & \textemdash \\
HiMatch (ACL2021) & 85.04$\pm$0.9 & 80.37$\pm$1.3 & 84.42$\pm$0.9 & 63.99$\pm$1.1 & 76.70$\pm$0.3 & 59.86$\pm$0.5 & 76.63$\pm$0.8 & 57.87$\pm$1.0 & \textemdash & \textemdash \\ \midrule
\multicolumn{11}{c}{\textbf{Pretrained Language Models}} \\ \hline
T5(2019) & 85.63$\pm$0.5 & 79.09$\pm$0.9 & 84.04$\pm$0.5 & 65.08$\pm$0.8 & 74.99$\pm$0.7 & 61.10$\pm$0.3 & 74.82$\pm$0.9 & 61.09$\pm$0.8 & \textemdash & \textemdash \\
RoBERTa (2019) & 84.76$\pm$0.8 & 79.11$\pm$0.3 & 83.45$\pm$0.2 & 65.01$\pm$0.9 & 75.08$\pm$0.6 & 60.31$\pm$0.8 & 74.16$\pm$0.9 & 60.49$\pm$1.0 & \textemdash & \textemdash \\
HGCLR (ACL2022) & 86.75$\pm$0.3 & 80.93$\pm$1.0 & 86.40$\pm$0.7 & 68.32$\pm$0.5 & 78.90$\pm$0.3 & 62.75$\pm$0.4 & 78.34$\pm$0.8 & 64.96$\pm$0.7 & \textemdash & \textemdash \\
HPT (EMNLP2022) & 86.83$\pm$1.1 & 81.24$\pm$0.6 & 86.51$\pm$1.2 & 69.16$\pm$0.9 & 77.83$\pm$0.5 & 62.94$\pm$0.6 & 79.98$\pm$1.0 & 65.30$\pm$0.4 & \textemdash & \textemdash \\
HiTIN (ACL2023) & 87.52$\pm$0.2 & 81.96$\pm$0.4 & 86.79$\pm$0.6 & 69.65$\pm$0.8 & 78.60$\pm$0.3 & 63.22$\pm$0.4 & 79.98$\pm$0.6 & 65.08$\pm$0.5 & \textemdash & \textemdash \\
HJCL (EMNLP2023) & 87.67$\pm$0.4 & 81.45$\pm$0.3 & 86.27$\pm$0.4 & 69.40$\pm$0.9 & 78.55$\pm$0.8 & 63.25$\pm$0.6 & 80.36$\pm$0.4 & 66.60$\pm$0.2 & \textemdash & \textemdash \\
T5-InterMRC (IPM2024) & 86.04$\pm$0.7 & 80.73$\pm$0.5 & 84.96$\pm$1.0 & 66.05$\pm$0.4 & 77.22$\pm$0.9 & 61.78$\pm$1.1 & 75.83$\pm$0.8 & 63.11$\pm$0.7 & 74.89$\pm$0.7 & 65.14$\pm$0.8 \\
HiAdv (COLING2024) & 87.05$\pm$0.2 & 80.97$\pm$1.0 & 86.71$\pm$0.2 & 69.29$\pm$0.4 & 78.94$\pm$0.3 & 62.52$\pm$0.8 & 79.60$\pm$0.9 & 65.63$\pm$0.8 & 79.05$\pm$0.3 & 69.13$\pm$0.5 \\
HALB (KBS2024) & 86.75$\pm$1.0 & 81.44$\pm$0.8 & 86.14$\pm$0.9 & 68.38$\pm$0.8 & 78.10$\pm$0.4 & 63.10$\pm$0.5 & 80.05$\pm$0.4 & 66.21$\pm$0.6 & 79.09$\pm$0.2 & 69.20$\pm$0.3 \\
HILL (NAACL2024) & 87.44$\pm$0.2 & 81.83$\pm$0.5 & 86.23$\pm$0.4 & 69.50$\pm$0.6 & 79.02$\pm$0.3 & 63.24$\pm$0.4 & 79.97$\pm$0.3 & 65.84$\pm$0.7 & 79.25$\pm$0.6 & 69.54$\pm$0.6 \\
HiGen (EACL2024) & 87.39$\pm$0.2 & 81.65$\pm$0.2 & 86.53$\pm$0.4 & 69.24$\pm$0.3 & 78.71$\pm$0.5 & 63.09$\pm$0.5 & 80.39$\pm$0.3 & 66.54$\pm$0.6 & 78.62$\pm$0.5 & 68.81$\pm$0.6 \\
\midrule
UME (Ours) & \textbf{87.71$\pm$0.4} & \textbf{82.29$\pm$0.3} & \textbf{86.88$\pm$0.4} & \textbf{69.70$\pm$0.2} & \textbf{79.21$\pm$0.2} & \textbf{63.46$\pm$0.3} & \textbf{80.46$\pm$0.6} & \textbf{66.74$\pm$0.4} & \textbf{79.41$\pm$0.3} & \textbf{69.76$\pm$0.5} \\
\midrule
\end{tabular}
\end{table*}

\begin{table}[htb]
    \caption{Parameter description and details}
    \label{tab:Parameter_description}
    \centering
    \begin{tabular}{cc}
    \hline
    Parameters & Description \\ \hline
    r          & LoRA Rank. Default: 64 \\
    lr         & Learning rate \\
    batch      & Batch size \\
    epoch      & Default: 10 \\
    early-stop & \makecell[c]{Epoch before early stop} \\
    update     & \makecell[c]{Gradient accumulate steps} \\
    warmup     & Warm-up steps \\
    graph      & \makecell[c]{Whether to use graph encoder. \\ Default: True} \\
    multi      & \makecell[c]{Whether the task is multi-label \\ classification. Should keep default \\ since all datasets are multi-label \\ classifications. Default: True } \\
    thre       & \makecell[c]{Threshold for keeping tokens. \\ Denoted as gamma in the paper} \\
    wandb      & Use wandb for logging \\
    ta         & \makecell[c]{If prefix weight $\leq \varepsilon$, the loss of \\ expert m on the sample will be eliminated} \\
    eta        & \makecell[c]{Eta is a temperature factor that adjusts \\ the sensitivity of prefix weights} \\ \hline
    \end{tabular}
\end{table}

\subsection{Baseline and Parameter Settings}  
We select 14 competitive baseline models for comparisons, including HiAGM \cite{2020HiAGM}, HTCInfoMax \cite{2021HTCinfomax}, HiMatch \cite{2021HiMatch}, T5 \cite{2020T5}, RoBERTa \cite{2019roberta}, HGCLR \cite{2022contrast}, HPT \cite{2022HPT}, HJCL \cite{2023HJCL}, HiTIN \cite{2023HiTIN}, T5-InterMRC \cite{2024t5-inter}, HiAdv \cite{2024HiAdv}, HALB \cite{2024HALB}, HILL \cite{2024HILL} and HiGen \cite{2024HiGen}. Specifically, HiAGM uses label-dependent prior probability to aggregate node information for global-level awareness. HTCInfoMAX addresses the issues by utilizing information maximization to optimize text-label relationships and label representations. HiMatch maps text and tags to a common embedded space for better matching. The above three models all belong to hierarchy-aware models and don't use pretrained language models. In addition, T5 unifies natural language tasks into a text-to-text format, simplifying the fine-tuning process. RoBERTa improves the performance of BERT by increasing the model size and enhancing training methods. T5-InterMRC proposes an interpretable model based on T5 that enhances the explainability of evidence in the task. HiGen optimizes HTC performance through dynamic text representation and level-oriented loss functions. As shown in Table \ref{tab:Parameter_description}, we present the detailed training parameters, including LoRA rank, the number of epochs, and other hyperparameters, along with their meanings. Additionally, we conduct experiments to reflect the impact of different LoRA rank sizes on performance. Furthermore, we specify the use of BERT as the backbone for text and Graphormer as the classification hierarchy. Please refer to our GitHub repository for the complete implementation and reproducibility details.

\subsection{Main Results on Hierarchical Text Classification (RQ1)} 
The main average results by five-fold cross-validation are summarized in Table \ref{tab:ALL_MODELS}. Particularly, UME achieves the highest performance across all datasets and overall averages. Compared to HiMatch, the model with the highest average performance among Hierarchy-Aware models without using pretrained models, UME achieves improvements of 3.83\% in Micro-F1 and 8.87\% in Macro-F1 on BGC. On one hand, UME leverages the rich semantic information from pretrained models, which enhances its generalization ability. On the other hand, the propagation of uncertainty across multiple experts helps in identifying challenging samples at the lower levels. Additionally, among the baselines, HJCL shows the best average performance. It uses supervised contrastive learning to provide more comprehensive training for classes with a larger number of samples. In contrast, UME focuses on fewer experts that handle all samples. When encountering challenging minority samples, it activates more experts for ensemble classification. As a result, UME achieves an average improvement of 0.36\% in Micro-F1 and 0.37\% in Macro-F1 across the four datasets. Overall, UME consistently outperforms other strong baselines in all F1-related evaluations, demonstrating the effectiveness of using multiple low-rank experts in hierarchical text classification (HTC). In Table \ref{tab:std}, we repeat the experiments for the proposed method and the baseline HiTIN five times (using different random seeds). We conduct experiments on the WOS and RCV1-V2 datasets. It can be seen that these improvements are statistically significant based on the paired t-test at the 95\% significance level.

\subsection{Results on Challenging-tailed Categories (RQ2)}
\begin{table}[t!]
    \caption{The performance of different models on challenging-tailed categories in the RCV1-V2 dataset. We report results regarding Win-Lose numbers.}
    \label{tab:long-tailed}
    \centering
    \scriptsize
    \begin{tabular}{ccccccc}
    \midrule
        Category & Count & Ours & HiTIN & HJCL & HGCLR & HPT\\ \midrule
        g157 & 1991 & \textBF{55.6}5\% & 55.25\%  & 51.28\% & 49.42\% & 50.77\% \\ 
        c16 & 1871 & \textBF{58.04}\% & 54.62\%  & 57.96\% & 53.82\% & 55.41\% \\ 
        gwelf & 1818 & 48.01\% & \textBF{51.65}\%  & 49.87\% & 45.21\% & 45.21\% \\ 
        e311 & 1658 & \textBF{79.79}\% & 61.46\%  & 58.74\% & 61.82\% & 59.18\% \\ 
        c331 & 1179 & 77.26\% & 67.77\%  & \textBF{79.55}\% & 54.20\% & 74.36\% \\ 
        e143 & 1172 & \textBF{77.39}\% & 65.61\%  & 66.20\% & 65.87\% & 65.31\% \\ 
        c313 & 1074 & 47.21\% & 42.64\%  & \textBF{49.27}\% & 39.11\% & 29.23\% \\ 
        e132 & 922 & 58.79\% & \textBF{64.97}\%  & 59.01\% & 57.92\% & 64.22\% \\ 
        gobit & 831 & \textBF{72.20}\% & 57.16\%  & 58.18\% & 47.65\% & 54.11\% \\ 
        gtour & 657 & 66.33\% & 59.06\%  & \textBF{67.06}\% & 54.49\% & 64.51\% \\ 
        e61 & 376 & \textBF{69.41}\% & 66.76\%  & 58.67\% & 64.10\% & 64.36\% \\ 
        e141 & 364 & 58.52\% & \textBF{61.54}\%  & 60.38\% & 39.56\% & 59.92\% \\ 
        gfas & 307 & \textBF{2.61}\% & 0.00\%  & 0.00\% & 0.00\% & 0.00\% \\ 
        e142 & 192 & 50.93\% & 46.35\%  & \textBF{51.25}\% & 46.88\% & 49.98\% \\ 
        e313 & 108 & \textBF{4.63}\% & 0.00\%  & 2.78\% & 0.00\% & 0.00\% \\ 
        e312 & 52 & \textBF{3.85}\% & 0.00\%  & 0.00\% & 0.00\% & 0.00\% \\ \midrule
        Result & Win-Lose & \textBF{9-7} & 3-13  & 4-12 & 0-16 & 0-16 \\ \midrule
    \end{tabular}
\end{table}

\begin{table}[htb]
    \caption{The Macro-F1 performance on tailed labels for the Top N least frequent labels across five datasets.}
    \label{tab:Top-N}
    \centering
    \scriptsize
    % Reduce column separation to try and make it fit
    \setlength{\tabcolsep}{3pt} 
    % This new line increases the vertical spacing between rows by 20%
    \renewcommand{\arraystretch}{1.2}
    \begin{tabular}{cccccccc}
    \hline
    \multicolumn{1}{l}{Datasets} & \begin{tabular}[c]{@{}c@{}}Top N \\ Tailed Labels\end{tabular} & HGCLR & HPT   & HiTIN & HJCL  & HILL  & Ours    \\ \hline
    \multirow{3}{*}{WOS}         & N=8                         & 28.47 & 27.78 & 37.83 & 41.67 & 42.08 & \textbf{46.60} \\
                                 & N=16                        & 48.67 & 49.04 & 53.16 & 57.67   & 51.78 & \textbf{62.66}  \\
                                 & N=32                        & 61.49 & 65.36 & 61.43 & 71.62 & 68.44 & \textbf{74.70}  \\ \hline
    \multirow{3}{*}{RCV1-V2}     & N=8                         & 18.92 & 16.88 & 20.24 & 22.40 & 18.38 & \textbf{26.93}  \\
                                 & N=16                        & 41.83 & 41.37 & 47.21 & 43.22 & 46.72 & \textbf{50.45}  \\
                                 & N=32                        & 56.67 & 54.77  & 57.98 & 57.20 & 61.46 & \textbf{63.69}  \\ \hline
    \multirow{3}{*}{AAPD}        & N=8                         & 28.35 & 36.50 & 38.95 & 37.60 & 40.46 & \textbf{41.71}  \\
                                 & N=16                        & 45.47 & 44.69 & 51.90 & 49.83 & 49.97 & \textbf{53.49}  \\
                                 & N=32                        & 54.76 & 54.82 & 57.69 & 56.59 & 57.63 & \textbf{59.11}  \\ \hline
    \multirow{3}{*}{BGC}         & N=8                         & 20.12 & 26.45 & 21.88 & 27.96 & 21.30 & \textbf{32.40}  \\
                                 & N=16                        & 45.83 & 45.33 & 41.68 & 37.89 & 36.27 & \textbf{46.78}  \\
                                 & N=32                        & 54.29 & 52.90 & 51.16 & 54.05 & 50.58 & \textbf{56.16}  \\ \hline
    \multirow{3}{*}{NYT}         & N=8                         & 9.38  & 12.83 & 13.52 & 12.82 & 14.31 & \textbf{15.36}  \\
                                 & N=16                        & 22.43 & 21.21 & 28.70 & 34.39 & 35.31 & \textbf{35.78}  \\
                                 & N=32                        & 48.36 & 45.21 & 47.39 & 50.53 & 50.23 & \textbf{50.75}  \\ \hline
    \end{tabular}
\end{table}

\begin{figure}[t!] % Changed from figure* to figure
    \centering
    \subfloat[HGCLR]{
        \includegraphics[width=0.23\textwidth]{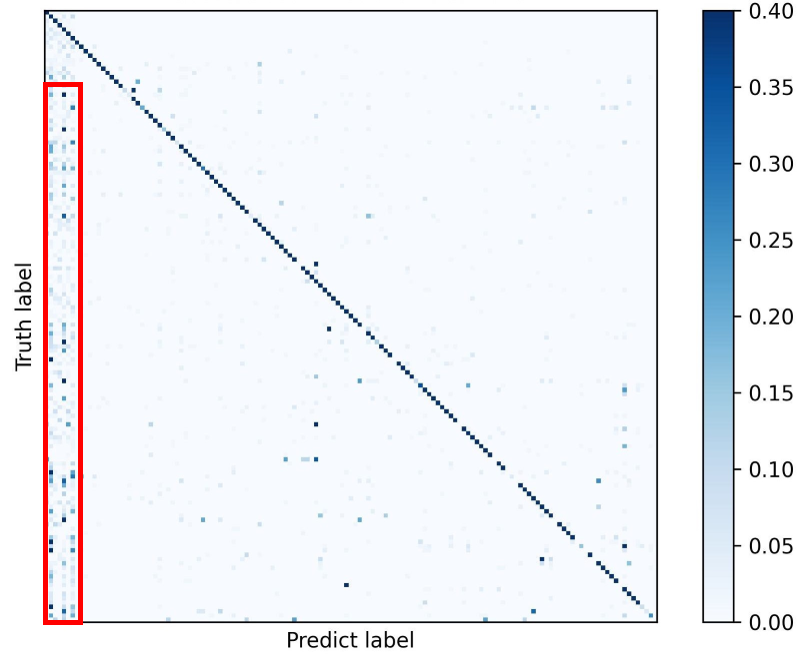} % Adjusted width for single column
    }
    \hfil
    \subfloat[UME]{
        \includegraphics[width=0.23\textwidth]{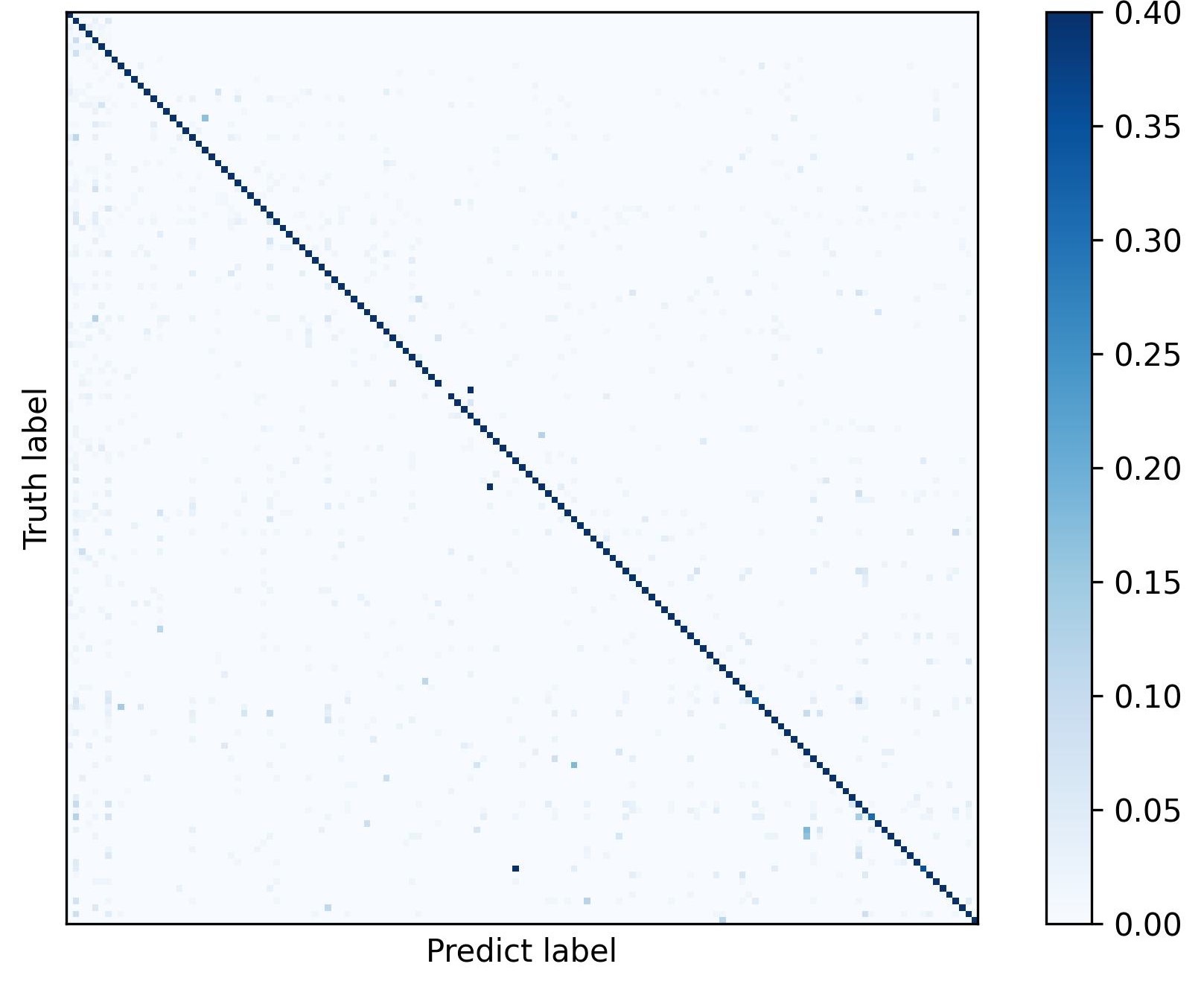} % Adjusted width for single column
    }
    \caption{The confusion matrix highlights the contrast between HGCLR and multi-experts in terms of prediction accuracy.}
    \label{fig:confusion}
\end{figure}

We discuss the performance of the models on those challenging-tailed categories. Specifically, we intentionally pick up the categories of less than 2000 samples in RCV1-V2 as shown in Table \ref{tab:long-tailed}. Compared with HiTIN, HJCL, HGCLR and HPT, our proposed UME outperforms them regarding the overall win-lose numbers, achieving the best performance in 9/16 challenging-tailed categories. In detail, for categories such as ``e311", ``gtour", and ``e61", UME achieves the best performance with much higher accuracy percentages than others. In the ``e311'' category, UME achieves an impressive accuracy of 79.79\%, surpassing the accuracy of others. Moreover, we find that all other baselines cannot correctly classify samples from categories of ``gfas" and ``e312" (all baselines' are 0's on these categories), UME still works in these categories. In addition, we test the Macro-F1 of the samples corresponding to the top N least frequent tail labels across five datasets, as shown in Table \ref{tab:Top-N}. Compared with HiTIN, HJCL, HGCLR, HPT, and HILL, we achieve state-of-the-art performance on the tail classes. On the WOS dataset, we achieve an average improvement of 11.03\%, 10.60\%, and 9.03\% on the Top 8, Top 16, and Top 32 least frequent tail labels, respectively. This demonstrates that the use of uncertainty-based multi-expert fusion can alleviate the issue of challenging-tailed label distributions in HTC tasks.

\begin{figure}[t!]
\centering
% The width is changed to the column's width
\includegraphics[width=\columnwidth]{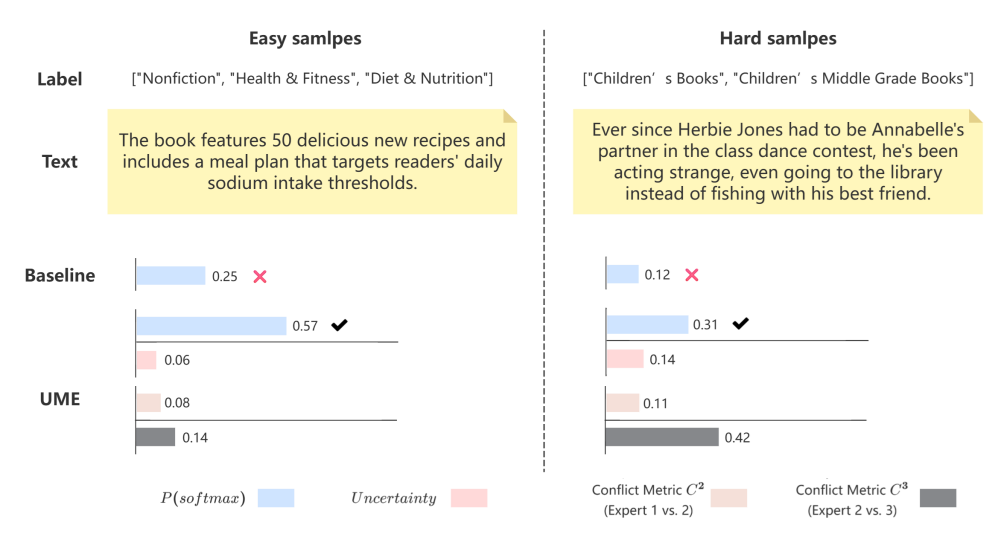}
\caption{Two examples illustrate the effectiveness of uncertainty-based multi-expert fusion. Compared with baselines, our classification results are more reliable.}
\label{fig:case_study}
\end{figure}

\begin{figure}[t!]
\centering
% The width is changed to the column's width
\includegraphics[width=\columnwidth]{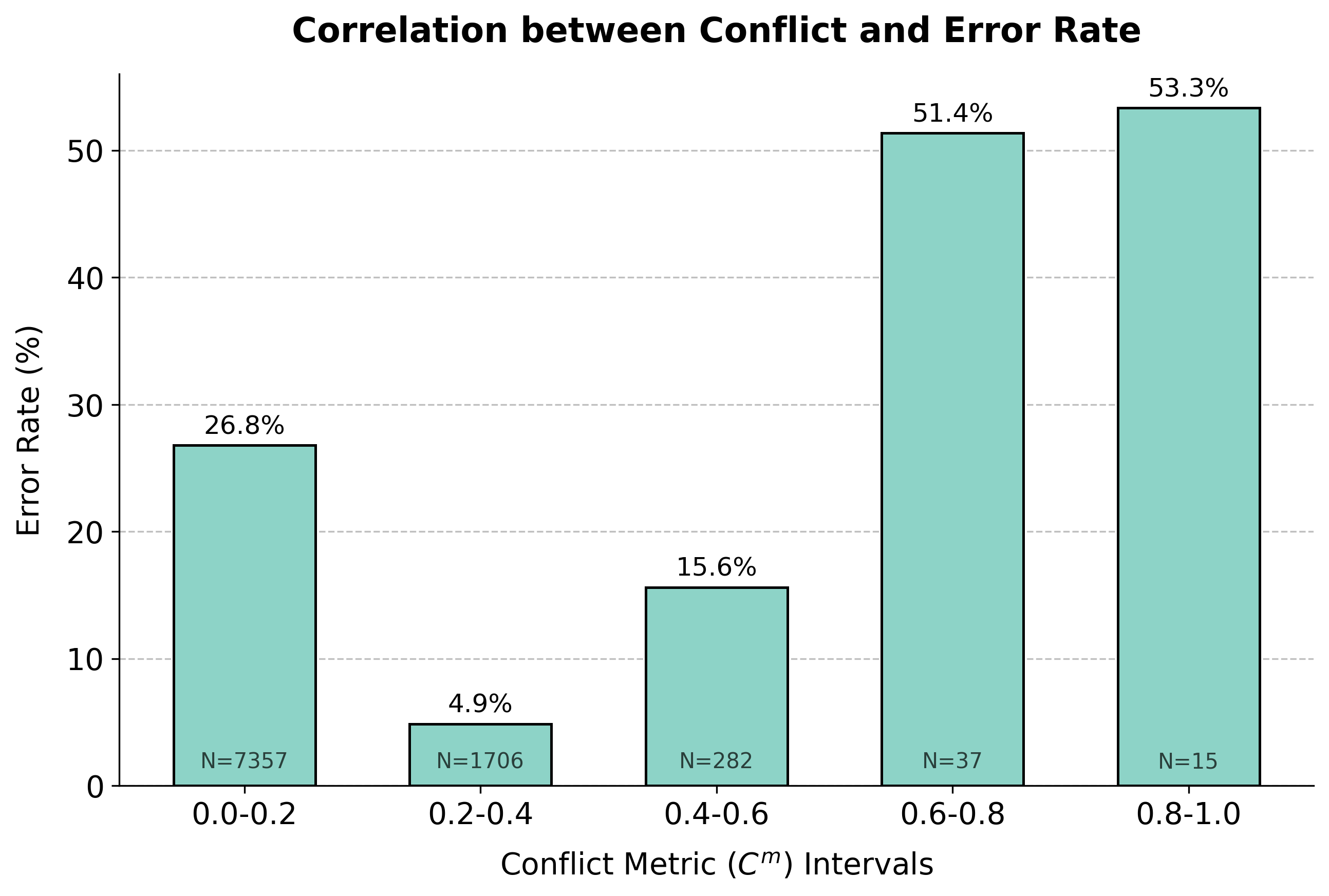}
\caption{The classification error rate distribution across different Conflict Metric ($C^m$) intervals. $N$ denotes the number of samples in each bin.}
\label{fig:conflict_error}
\end{figure}

In Figure \ref{fig:case_study}, we present two real examples from the BGC dataset. One is a simple sample and the other is a challenging sample. Compared to the baseline, UME demonstrates higher softmax classification probabilities and produces the correct prediction results. We further illustrate the reliability of the proposed UME by applying Dempster Shafer Evidence Theory. This framework helps multiple experts reach accurate uncertainty estimates. We specifically analyzed the Conflict Metrics to understand expert disagreement. For the simple sample, the conflict scores $C^2$ and $C^3$ remain low at 0.08 and 0.14 respectively. This indicates a high consensus among experts. In contrast, the challenging sample exhibits a slight increase in $C^2$ to 0.11 and a sharp spike in $C^3$ to 0.42. This significant disagreement correctly signals the semantic ambiguity of the text. The fusion mechanism effectively handles this conflict to output confident and reliable classification results.

As illustrated by our binning analysis in Figure \ref{fig:conflict_error}, the classification error rate exhibits a strong positive correlation with the conflict metric $C^m$ beyond the initial interval, where the error rate sharply escalates from 4.9\% in low-conflict regions to over 50\% in high-conflict buckets ($C^m > 0.6$). The relatively higher error rate in the 0.0-0.2 range (26.8\%) reflects a confidently wrong phenomenon, where experts reach a unanimous but incorrect consensus due to shared biases or misleading features. This overall performance trend empirically validates $C^m$ as a reliable indicator of prediction risk and justifies our use of Dempster-Shafer Theory. By identifying these challenging samples where expert evidence diverges—typically occurring in fine-grained or long-tailed categories—the UME framework can dynamically resolve conflicts and enhance decision reliability through its uncertainty-guided fusion mechanism.

\begin{figure}[t!]
    \centering
    \subfloat[WOS]{
        \includegraphics[width=0.47\linewidth]{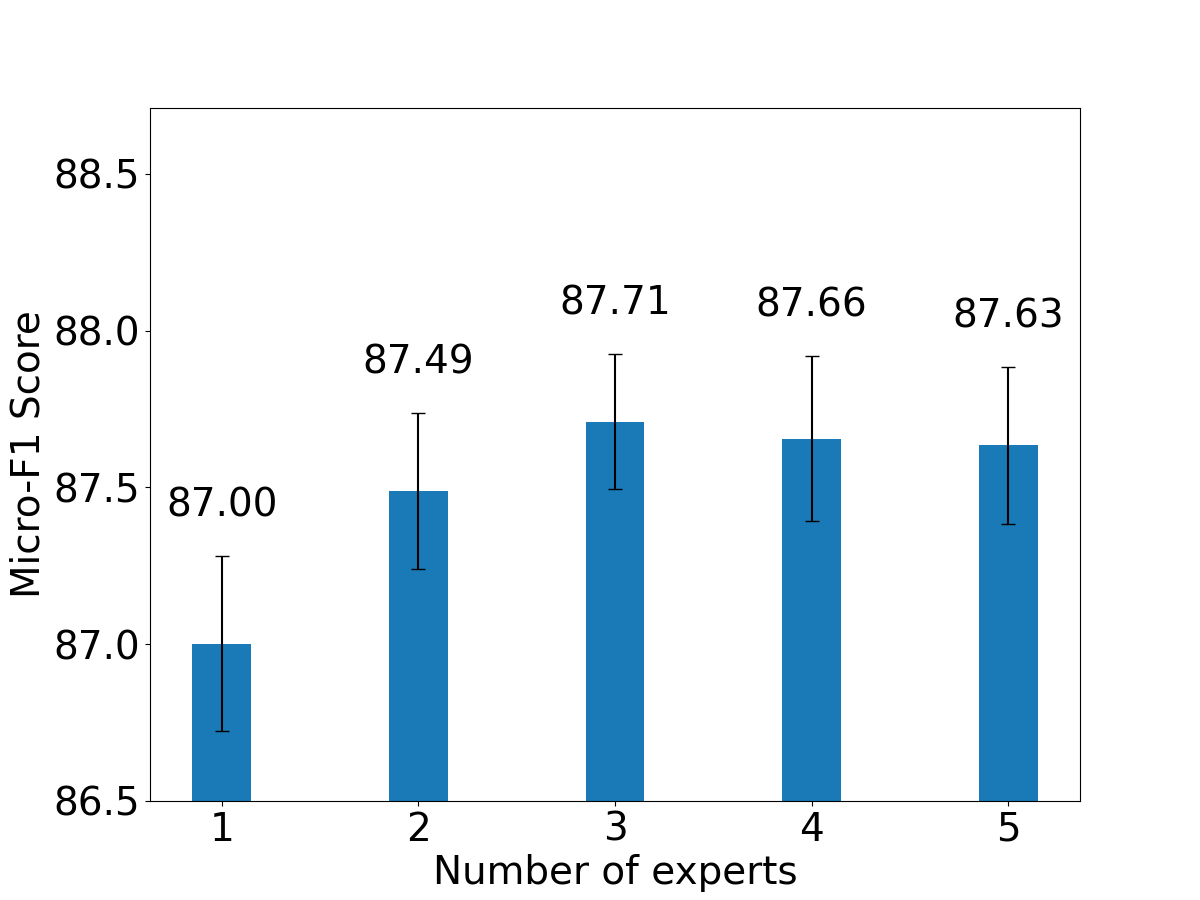}
    }
    \hfil
    \subfloat[RCV1-V2]{
        \includegraphics[width=0.47\linewidth]{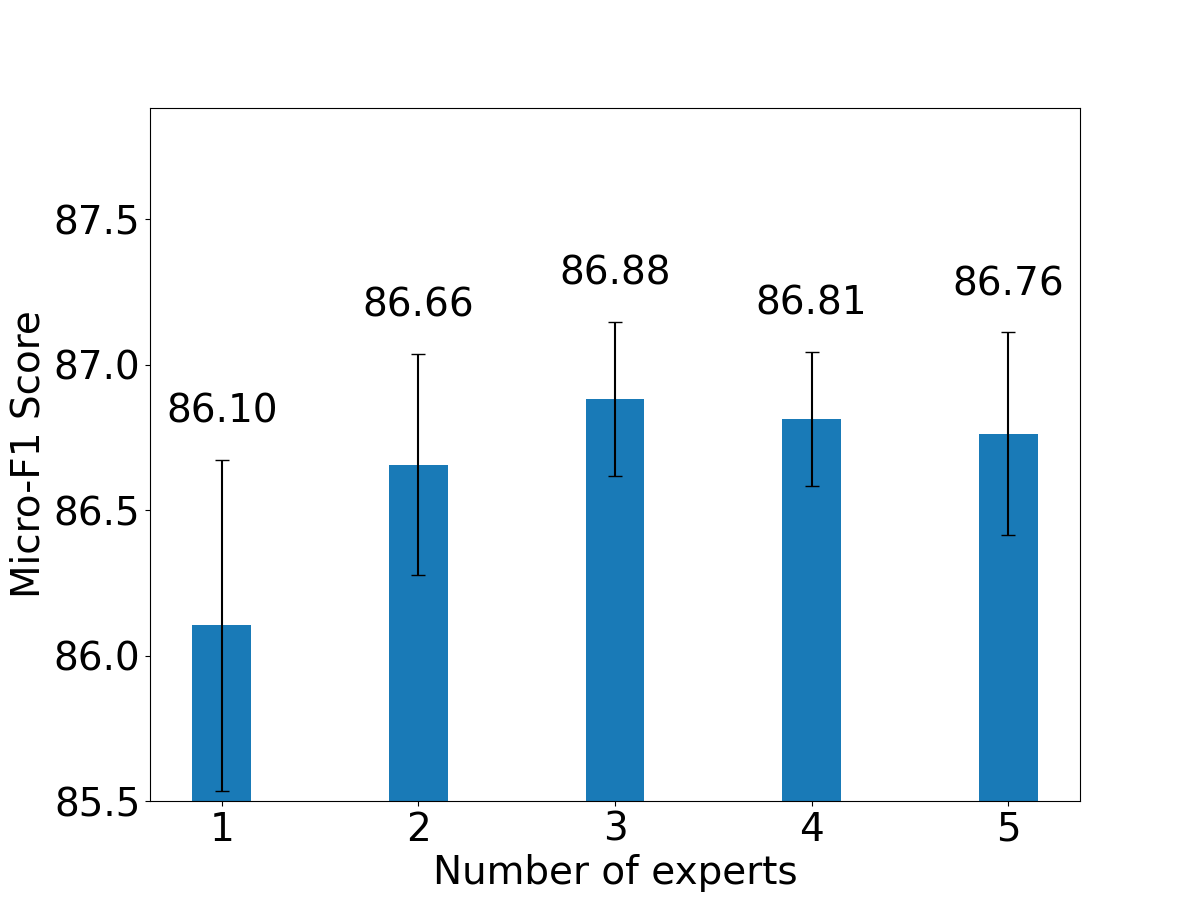}
    }
    \\ % Creates a new line for the next row of images
    \subfloat[AAPD]{
        \includegraphics[width=0.47\linewidth]{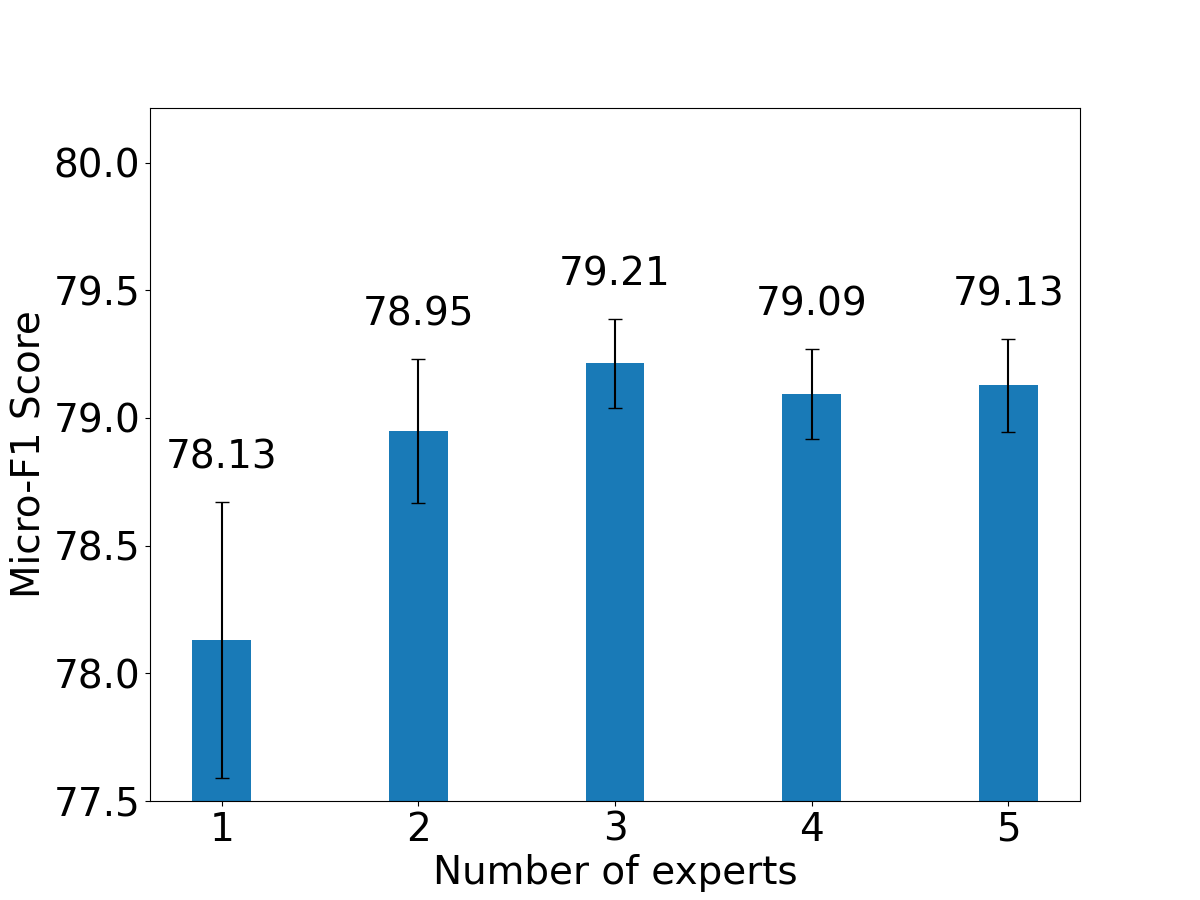}
    }
    \hfil
    \subfloat[BGC]{
        \includegraphics[width=0.47\linewidth]{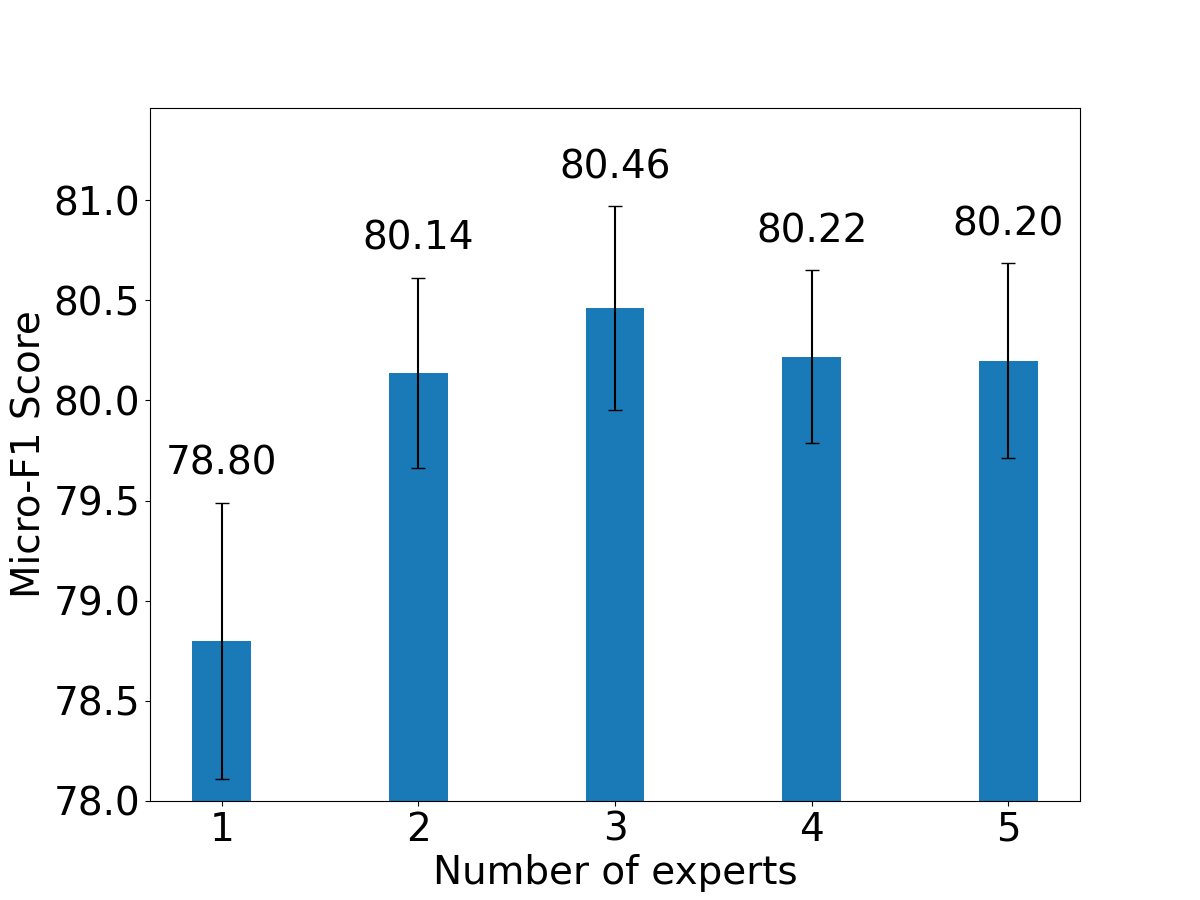}
    }
    \caption{Bar charts show the different performances with varying numbers of experts among four datasets.}
    \label{fig:std}
\end{figure}

\begin{table}[t]
\centering
\caption{Analysis of expert capacity and utilization on the WOS dataset with expert numbers ($M$). The metrics include Tail Micro-F1 (computed on the bottom 20\% least frequent classes), the Average Conflict of the final expert ($C^{last}$), and the proportion of samples routed to the final expert (defined as weight $>0.5$).}
\label{tab:expert_analysis}
\setlength{\tabcolsep}{3pt}
\renewcommand{\arraystretch}{1.2}
\begin{tabular}{c c c c}
\toprule
\ Experts ($M$) & Tail Micro-F1 & Avg. Conflict ($C^{last}$) & Last Expert Util. \\
\midrule
\textbf{3} & \textbf{76.21} & \textbf{0.1724} & \textbf{20.26} \\
4 & 75.32 & 0.1597 & 1.80 \\
5 & 74.05 & 0.1175 & 1.28 \\
\bottomrule
\end{tabular}
\end{table}

\subsection{Effects of the Number of Experts? (RQ3)} 
We further observe that increasing the number of experts leads to higher average conflict and fewer samples routed to later experts. In Figure \ref{fig:std}, we present an analysis with different numbers of experts. Clearly, as the number of experts increases, the Micro-F1 score performance initially rises to its highest with three experts and then gradually decreases. Furthermore, the observed improvements are consistent across multiple runs, as reflected by the reported standard deviations, suggesting these differences represent meaningful shifts in predictive capability rather than random chance. We attribute the peak at three experts to our Sequential Specialization strategy : since the datasets possess a hierarchical depth of only 2 to 4 levels, three experts naturally align with the Head, Medium, and Tail categories, effectively saturating the model's capacity to capture valid semantic distinctions. Consequently, adding experts beyond this point forces the model to overfit to residual noise or ambiguous outliers. These redundant experts introduce high conflict values during the DST fusion process, which slightly degrades global inference quality rather than providing further benefit.

As illustrated in Table \ref{tab:expert_analysis}, the model achieves optimal performance at $M=3$. To investigate the performance degradation at $M>3$, we analyzed the \textit{Last Expert Utilization} (the proportion of samples where the final expert's weight exceeds 0.5) and the \textit{Average Conflict Metric} ($C^{last}$), which quantifies the prediction discrepancy between the final expert and its predecessor.

At $M=3$, the final expert is utilized for 20.26 of the samples, maintaining a significant conflict signal between the second and third experts ($C^{last}=0.1724$). This indicates that the final expert receives sufficient discordant samples to learn effective representations for tail classes, resulting in the highest Tail Micro-F1 of 76.21. However, as $M$ increases to 4 and 5, the utilization rate of the final expert drops precipitously to 1.80 and 1.28, respectively. Simultaneously, the conflict between the last two experts diminishes to 0.1175. The preceding experts have already resolved the majority of semantic ambiguities, leaving the final expert with minimal conflict signals and insufficient training data. Consequently, the additional experts fail to capture meaningful patterns and instead introduce parameter noise, leading to a decline in tail performance. Thus, $M=3$ represents the optimal trade-off between expert specialization and data sufficiency.

\subsection{Experts Participation Percentages (RQ4)} 
\begin{figure}[t!]
\centering 
\includegraphics[width=0.5\textwidth]{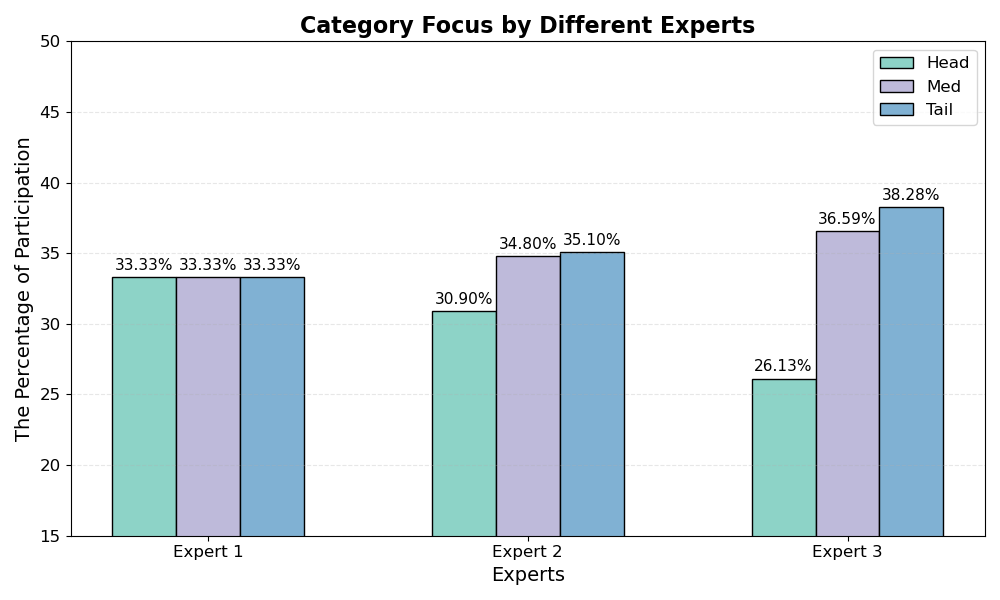}
\caption{The participation percentage among experts in the head, medium, and tail categories on RCV1-V2.}
\label{fig:preference}
\end{figure}

To analyze the involvement of each expert in the proposed model, we visualize the percentage of dynamical weights $\{w^m\}$ assigned to different experts in Equation (\ref{eq:expert_weight}). These weights are used to predict samples from the head, medium, and tail of the hierarchy classes in the RCV1-V2 dataset. The head, medium, and tail samples correspond to the 1st, 2nd, and 3rd level categories, respectively. Figure \ref{fig:preference} illustrates that the initial experts exhibit relatively balanced participation across the classes. However, as we gradually increase the number of experts, they tend to focus more on the challenging-tailed samples at the bottom level. This observation aligns with our initial design concept, where uncertainty is propagated from preceding experts to later experts, leading to increased attention on the challenging-tailed samples.

\begin{figure}[t]
    \centering
    % 第一个子图
    \subfloat[Impact of temperature factor $\eta$]{
        \includegraphics[width=0.2\textwidth]{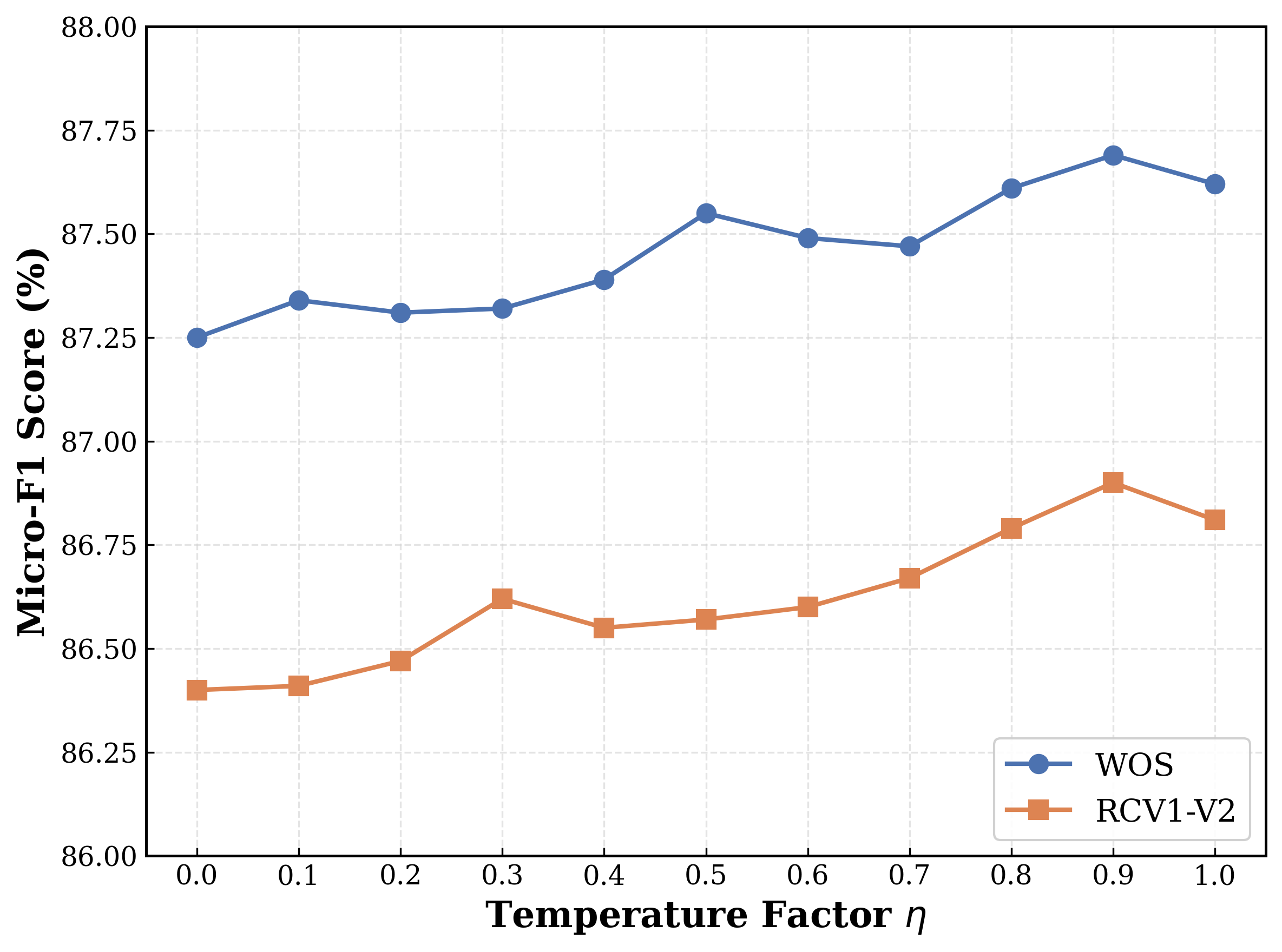}
        \label{fig:temperature_sensitivity}
    }
    \hfill % 左右间距
    % 第二个子图
    \subfloat[Impact of conflict metric threshold $\epsilon$]{
        \includegraphics[width=0.2\textwidth]{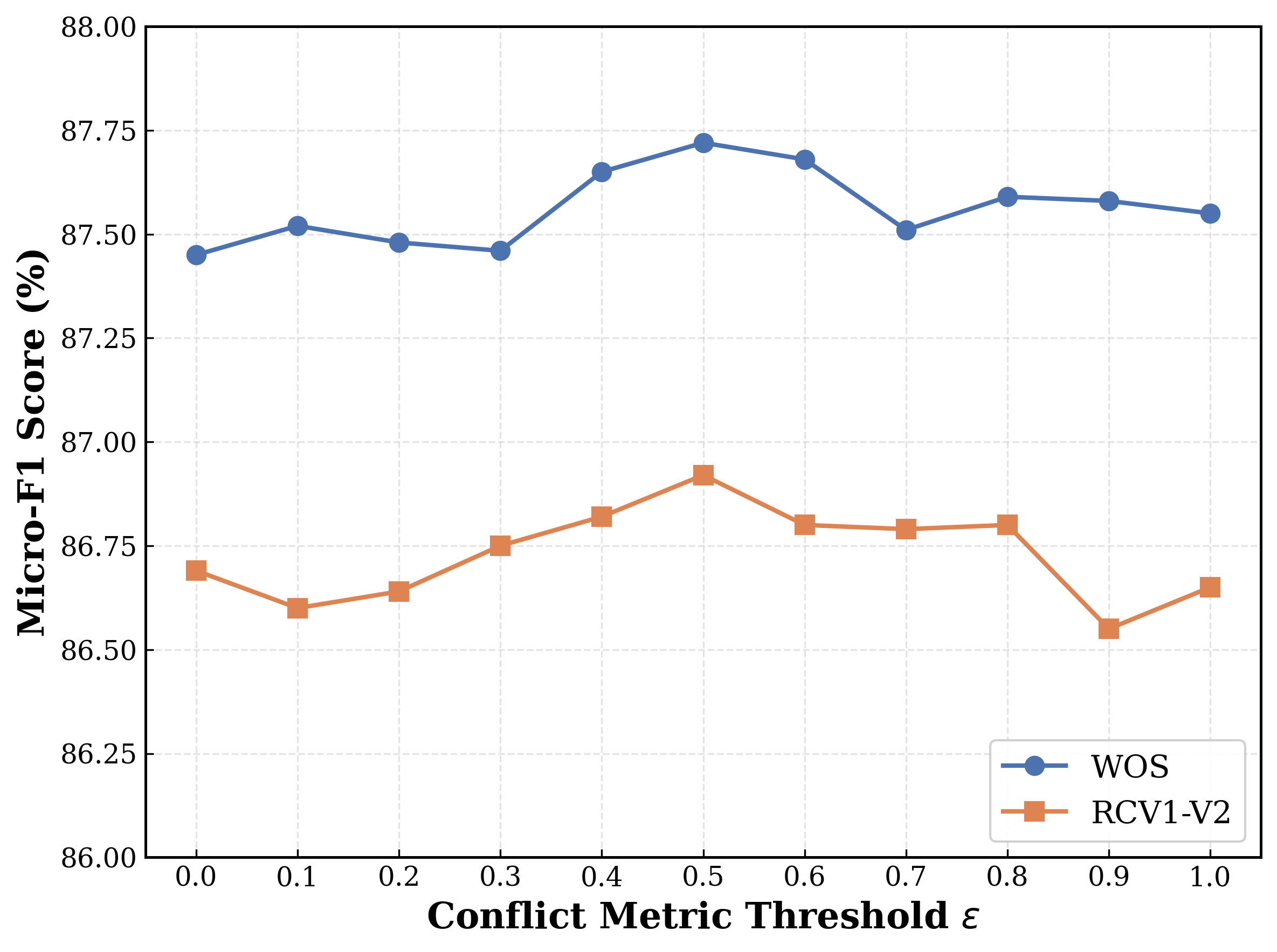}
        \label{fig:conflict_sensitivity}
    }
    
    \caption{Sensitivity analysis on WOS and RCV1-V2.}
    \label{fig:sensitivity_analysis}
\end{figure}

\subsection{Component Analysis (RQ5)}
We first analyze the contribution of each component. Specifically, we compare the full UME model against three variants. These include: i) without $\mathcal{L}_{evidence}$; ii) without ${{\mathcal{L}}_{evidence}}\And {{\mathcal{L}}_{ml}}$; and iii) without ${{\mathcal{L}}_{evidence}}\And {{\mathcal{L}}_{ml}}\And \widehat{{{\mathcal{L}}^{\mathcal{C}}}}\And {{\mathcal{L}}^{cl}}$. Ablation results are summarized in Table \ref{tab:ablation}. As shown, each component plays a distinct role. The $\widehat{{{\mathcal{L}}^{\mathcal{C}}}}\And {{\mathcal{L}}^{cl}}$ component is the most influential because it integrates hierarchical structure information. Additionally, the $\mathcal{L}_{ml}$ component significantly bolsters the model. It leverages prior evidence to delineate objectives through marginal likelihood. Conversely, $\mathcal{L}_{evidence}$ yields a modest impact. It improves performance by diminishing the evidence of incorrect labels and instilling greater uncertainty in erroneous classifications. Crucially, we verified whether the gains stem from our specific fusion design or merely from ensemble capacity. 

We further verify whether the performance gains stem from uncertainty-aware fusion rather than ensemble capacity. The first is Direct Integration, which simply averages expert outputs. The second is Gated MoE, which employs a learnable gating network for expert routing. As shown in the bottom section of Table \ref{tab:ablation}, both methods consistently underperform compared to UME across all datasets. For example, Gated MoE achieves 86.74 on WOS versus 87.71 for UME. This decline indicates that standard routing mechanisms or simple averaging fail to coordinate sequentially trained experts effectively. They struggle to account for the specific uncertainty patterns of residual experts handling hard tail samples. In contrast, UME's DST guided fusion dynamically prioritizes experts based on confidence. This indicates that the uncertainty aware mechanism is the key driver of performance, not just the number of experts.

To evaluate the robustness of UME, we conduct a comprehensive sensitivity analysis on the temperature factor $\eta$ and the conflict metric threshold $\epsilon$, which regulate the trade-off between early confident decisions and deeper expert involvement. As illustrated by our experiments in Figure \ref{fig:temperature_sensitivity}, the temperature factor $\eta$ modulates the sharpness of the fusion weights, achieving peak performance at $\eta = 0.9$ with Micro-F1 scores of 87.69 on WOS and 86.90 on RCV1-V2, while maintaining high stability across the entire interval. Similarly, the conflict threshold $\epsilon$ analyzed in Figure \ref{fig:conflict_sensitivity} determines when accumulated uncertainty triggers the participation of subsequent experts through our Sequential Specialization mechanism. Optimal results for this parameter are observed at $\epsilon = 0.5$, yielding peak scores of 87.72 for WOS and 86.92 for RCV1-V2, with minimal performance fluctuations across tested ranges. These findings empirically validate that the conflict metric $C^m$ serves as a stable indicator for expert coordination, ensuring that UME achieves state-of-the-art results without the need for exhaustive fine-tuning.

\begin{table}[t]
\caption{Ablation study results for Micro-F1.}
\label{tab:ablation}
\centering
\scriptsize
% No need to set \tabcolsep manually, tabularx handles spacing
% \setlength{\tabcolsep}{5pt} 
\begin{tabularx}{\columnwidth}{@{} l *{4}{>{\centering\arraybackslash}X} @{}}
\toprule
Ablation Study & WOS & RCV1-V2 & AAPD & BGC \\
\midrule
w/o $\mathcal{L}_{evidence}$ & 87.52 & 86.61 & 78.86 & 79.66 \\
w/o ($\mathcal{L}_{evidence}\&\mathcal{L}_{ml}$) & 86.98 & 86.37 & 78.15 & 79.27 \\
\makecell[l]{w/o ($\mathcal{L}_{evidence}\&\mathcal{L}_{ml}$\\$\& {\hat {\mathcal{L^C}}}\&{\mathcal{L}^{cl}}$)} & 85.91 & 85.72 & 77.28 & 78.51 \\
$\mbox{-}r.p.$ Direct Integration & 87.18 & 85.73 & 78.53 & 79.51 \\
$\mbox{-}r.p.$ Gated MoE & 86.74 & 86.12 & 77.91 & 79.08 \\
\midrule
\textbf{UME} & \textbf{87.71} & \textbf{86.88} & \textbf{79.21} & \textbf{80.46} \\
\bottomrule
\end{tabularx}
\end{table}

\begin{figure}[t!]
\centering 
\includegraphics[width=0.475\textwidth]{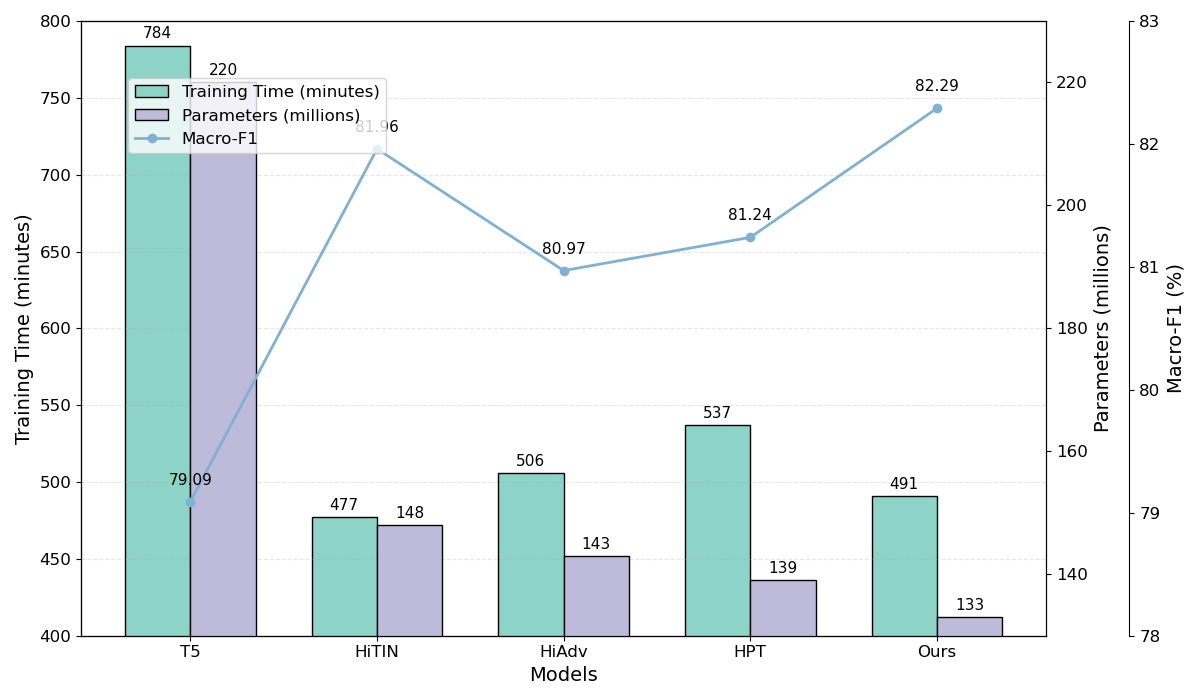}
\caption{Compare model parameters, training time and Macro-F1 scores on the WOS dataset.}
\label{fig:parameters}
\end{figure}

\subsection{Computational Costs (RQ6)}
In this section, we report the number of parameters, Macro-F1 scores, and classification accuracies of UME compared to T5 \cite{2020T5}, HPT \cite{2022HPT}, HiTIN \cite{2023HiTIN}, and HiAdv \cite{2024HiAdv} on the RCV1-V2 dataset. As shown in Figure \ref{fig:parameters}, we show the number of trainable parameters and training time for the model with three experts, demonstrating that this configuration achieves the best and most stable performance. UME contains fewer parameters than the other models but achieves the best performance. This is mainly because we modify only the FFN layer of the BERT encoder to design a multi-expert architecture. Furthermore, we use LoRA to freeze most of the expert parameters and fine-tune the multi-expert model with low-rank matrices. Compared to T5, UME reduces parameters by 39.49\% while improving the Macro-F1 score by 3.20\%. Parameter explosion in multi-expert fusion networks typically prevents the direct fusion of expert networks. To address this, we employ low-rank experts to improve efficiency, increasing parameters by only 1.18\% for each additional expert while achieving competitive performance. For total training time, UME is shorter than all models except HiTIN. Although our training time increases by 2.78\%, we reduce parameters by 10.37\% while achieving higher performance. Additionally, we measure inference speed and find no significant differences between models.

\begin{table}[t]
\centering
\caption{Performance comparison between LLMs and our UME framework on WOS and RCV1-V2 datasets. The LLMs are evaluated using the zero-shot prompt templates shown.}
\label{tab:llm_performance}
\setlength{\tabcolsep}{4pt}
\renewcommand{\arraystretch}{1.2}
\begin{tabular}{l c c c c}
\toprule
\multirow{2}{*}{Model} & \multicolumn{2}{c}{WOS} & \multicolumn{2}{c}{RCV1-V2} \\
\cmidrule(lr){2-3} \cmidrule(lr){4-5}
 & Micro-F1 & Macro-F1 & Micro-F1 & Macro-F1 \\
\midrule
GPT-4 & 59.74 & 37.80 & 56.12 & 36.54 \\
LlaMa-3 & 61.91 & 27.33 & 54.67 & 28.50 \\
UME (Ours) & \textbf{87.71} & \textbf{82.29} & \textbf{86.88} & \textbf{69.70} \\
\bottomrule
\end{tabular}
\end{table}

\begin{table}[t]
\caption{Prompt template for hierarchical text classification.}
\label{tab:prompt_template}
\centering
% This command increases row height for better vertical spacing, matching your example.
\renewcommand{\arraystretch}{1.2}
% m{...} 表示垂直居中且固定宽度的列 (需 array 包)
% 宽度设置为 \linewidth 的比例，确保适应单栏宽度
\begin{tabular}{ m{0.25\linewidth} m{0.6\linewidth} }
\hline
\centering \textbf{Prompt Template} & 
You are an expert hierarchical text classifier. Your task is to classify the given text into the correct category path from the predefined taxonomy. \newline
\newline
\textbf{[Taxonomy Definition]} \newline
The categories are defined in a hierarchy formatted as: $Level1 > Level2 > Level3$ \newline
- The symbol ``$>$'' represents the parent-child relationship. \newline
- You must output the full path from the root to the specific leaf category. \newline
\newline
\textbf{[Allowed Category Paths]} \newline
(Insert your full list of category paths here, e.g., Technology $>$ AI $>$ NLP) \newline
\newline
\textbf{[Constraints]} \newline
1. Output ONLY the category path. No explanation, no preamble. \newline
2. If the text fits multiple paths, output the most specific and relevant one. \newline
3. Use the exact string format provided in the [Allowed Category Paths]. \newline
\newline
\textbf{[Input Text]} \newline
\{\{text\_content\}\} \newline
\newline
\textbf{[Prediction]} \\ \hline
\end{tabular}
\end{table}

\subsection{Comparison with LLMs (RQ7)}
We further investigate the performance of Large Language Models (LLMs) on the HTC task to evaluate their zero-shot reasoning capabilities against our supervised approach. Table \ref{tab:llm_performance} presents the results of GPT-4 and Llama3, evaluated using the standardized prompt templates detailed in Table \ref{tab:prompt_template}.

As shown in Table \ref{tab:llm_performance}, GPT-4 achieves a Micro-F1 of 59.74 and Macro-F1 of 37.80 on the WOS dataset, with similar results on RCV1-V2 where it records 56.12 Micro-F1 and 36.54 Macro-F1. Llama3 demonstrates comparable performance, recording 61.91 Micro-F1 and 27.33 Macro-F1 on WOS, and 54.67 Micro-F1 and 28.50 Macro-F1 on RCV1-V2. These results indicate that instruction-tuned LLMs still significantly underperform compared to supervised fine-tuning models like UME, as UME achieves a Micro-F1 score exceeding 87 on WOS. The primary limitation is that the general reasoning capabilities of LLMs struggle to fully comprehend and align with the complex, structured dependencies of hierarchical label trees. They often fail to strictly adhere to the parent-child constraints required in HTC. Consequently, our proposed UME framework, which explicitly models hierarchy and uncertainty, proves to be both necessary and highly effective for achieving state-of-the-art precision.

\section{Conclusion}
%This paper proposed the Uncertainty-based Multi-Expert fusion network (UME) to address data imbalance in sequence learning. By integrating Ensemble LoRA, Sequential Specialization, and Uncertainty-Guided Fusion under the Dempster–Shafer Theory, UME enhances expert diversity and reliability with minimal parameter overhead. Experimental results on four hierarchical text classification benchmarks show that UME outperforms strong baselines, improving minority-class accuracy by up to 17.97\% while reducing trainable parameters by 10.32\%. These findings demonstrate that uncertainty-guided coordination among lightweight experts offers an effective solution for challenging-tailed sequence learning.
This work proposes the Uncertainty-based Multi-Expert fusion network (UME) for challenging-tailed sequence learning. By integrating Ensemble LoRA, Sequential Specialization, and Uncertainty-Guided Fusion under Dempster–Shafer Theory, UME improves tail-class performance while maintaining a compact parameter footprint. Although sequential specialization introduces a modest increase in training time due to reduced parallelism, this overhead represents a deliberate efficiency–performance trade-off that enables experts to focus on progressively harder samples. Experimental results demonstrate that this design yields substantial gains on minority classes with reduced trainable parameters.

In future work, we plan to further optimize training efficiency by exploring dynamic expert selection strategies that selectively activate experts based on uncertainty. Such mechanisms may alleviate sequential latency while preserving the effectiveness of uncertainty-guided specialization in resource-constrained scenarios.

\bibliographystyle{IEEEtran}
\bibliography{elsarticle}

\begin{IEEEbiography}[{\includegraphics[width=1in,height=1.25in,clip,keepaspectratio]{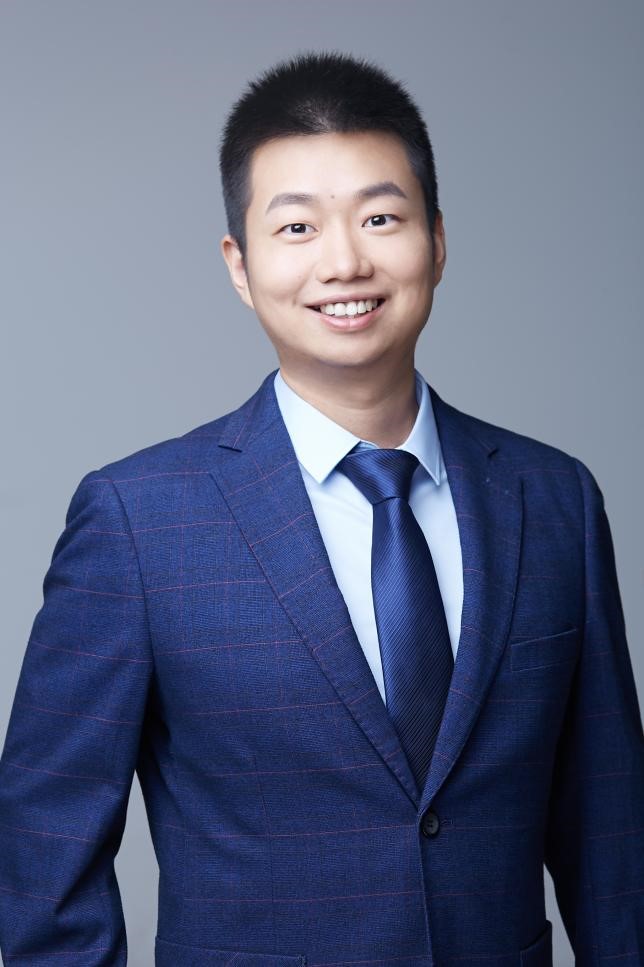}}]{Ye Wang}
Ye Wang received the B.S. degree in
Microelectronics from Chongqing University of Posts and Telecommunications,
Chongqing, China in 2011, the M.S. degree in Electrical Engineering from the
University of Texas at Dallas, Richardson, TX, USA in 2014, and the Ph.D.
degree in Computer Engineering from Texas A\&M University, College Station,
TX, USA in 2019. He joined the School of Artificial Intelligence at Chongqing
University of Posts and Telecommunications since 2020. His current research is
mainly focused on natural language processing and computer vision.
\end{IEEEbiography}

\begin{IEEEbiography}[{\includegraphics[width=1in,height=1.25in,clip,keepaspectratio]{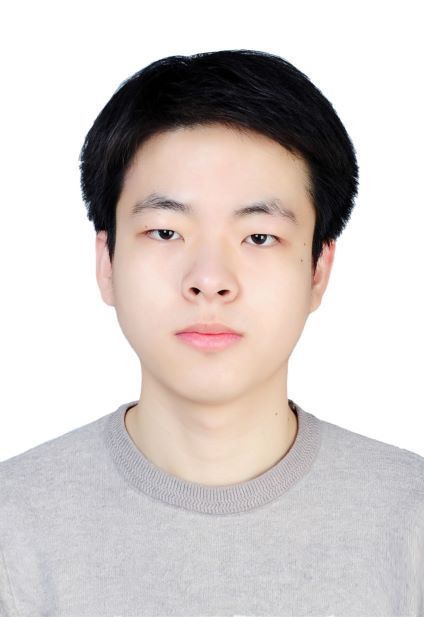}}]{Zixuan Wu}
Zixuan Wu received the B.S. degree in Information and Computing Science from Chongqing University of Posts and Telecommunications, Chongqing, China in 2022, and is pursuing the M.S. degree in Computer Science and Technology from the same university, with an expected graduation in 2025. His research is mainly focused on natural language processing and autonomous driving.
\end{IEEEbiography}

\begin{IEEEbiography}[{\includegraphics[width=1in,height=1.25in,clip,keepaspectratio]{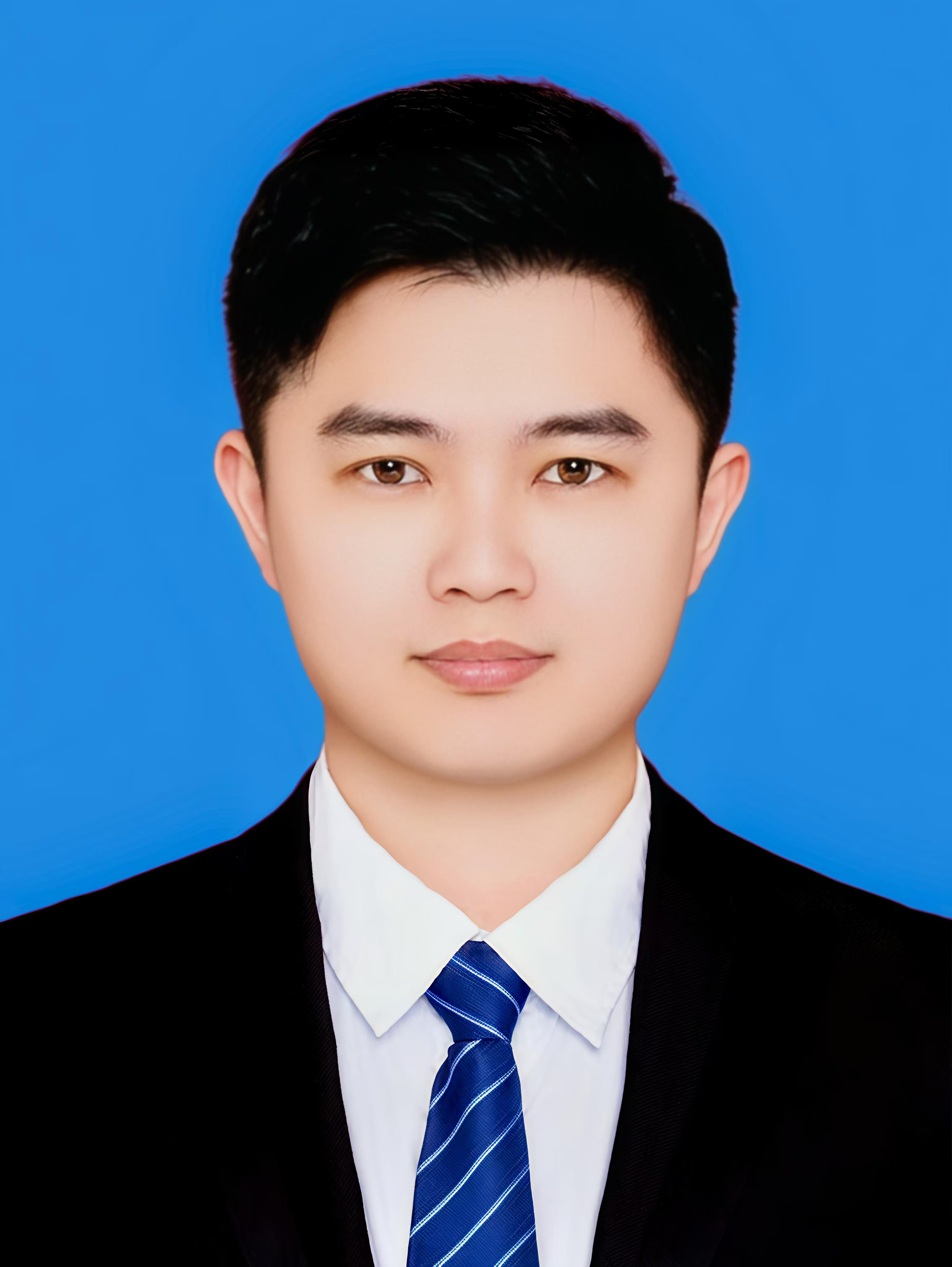}}]{Lifeng Shen}
Lifeng Shen (Member, IEEE) received the PhD degree in Artificial Intelligence from the Hong Kong University of Science and Technology, in 2024. He is an associate professor with the Department of Computer Science and Engineering, Chongqing University of Posts and Telecommunications (CQUPT). His current research interests include machine learning, deep learning, granular computing and time-series modeling and their applications.
\end{IEEEbiography}

\begin{IEEEbiography}[{\includegraphics[width=1in,height=1.25in,clip,keepaspectratio]{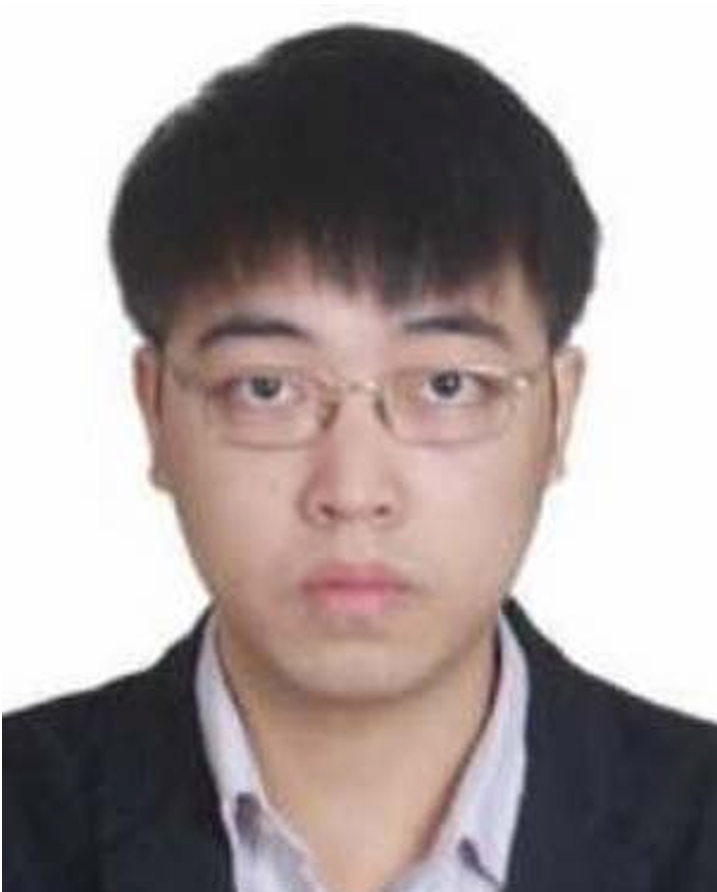}}]{Jiang Xie}
received the MS and PhD degrees in com puter science from Chongqing University, in 2015 and 2019, respectively. He is currently a lecturer with the College of Computer Science and Technology at Chongqing University of Posts and Telecommu nications. His research interests include clustering analysis and data mining.
\end{IEEEbiography}

\begin{IEEEbiography}[{\includegraphics[width=1in,height=1.25in,clip,keepaspectratio]{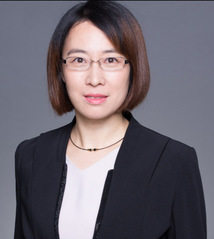}}]{Xiaoling Wang}
 (Member, IEEE) received the B.E.,
 M.S. and Ph.D. degrees from Southeast University in
 1997, 2000 and 2003, respectively. She is currently
 a Professor with the School of Computer Science at
 East China Normal University. Her research interests
 mainly include graph data processing and intelligent
 data analysis.
\end{IEEEbiography}

\vspace{-8mm}

\begin{IEEEbiography}[{\includegraphics[width=1in,height=1.25in,clip,keepaspectratio]{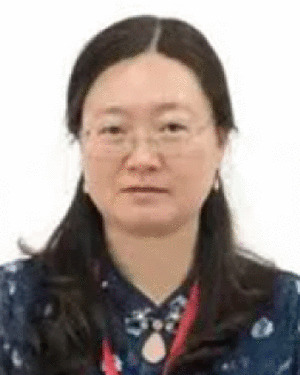}}]{Hong Yu}
Hong Yu (Member, IEEE) received the BE degree in physics from Nanchang Hangkong University, Nanchang, China, in 1994, the MS degree in signal and information processing from the Chongqing University of Posts and Telecommunications, Chongqing, China, in 1997, and the PhD degree in computer software and theory from Chongqing University, Chongqing, China, in 2003. She worked with the University of Regina, Canada, as a visiting scholar during 2007–2008. She is currently a full professor with the Chongqing University of Posts and Telecommunications, Chongqing, China. She received a Chongqing Natural Science and Technology Award. She has published several books and some peer-reviewed research articles. Her paper was selected as one of Top Articles in Outstanding S\&T Journal of China. Her research interests include rough sets, industrial Big Data, data mining, knowledge discovery, granular computing, three-way clustering, three-way decisions and intelligent recommendation.
\end{IEEEbiography}

\begin{IEEEbiography}
[{\includegraphics[width=1in,height=1.25in,clip,keepaspectratio]{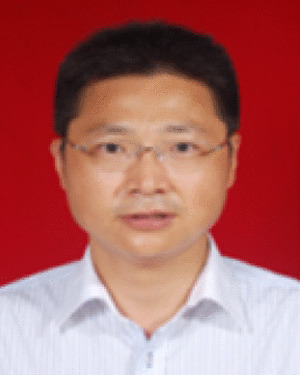}}]{Guoyin Wang}
Guoyin Wang (Senior Member, IEEE) received the BS, MS, and PhD degrees from Xi’an Jiaotong University, Xian, China, in 1992, 1994, and 1996, respectively. He worked with the University of North Texas, and the University of Regina, Canada, as a visiting scholar during 1998–1999. He had worked with the Chongqing University of Posts and Telecommunications during 1996–2024, where he was a professor, the vice-president with the University, the director of the Chongqing Key Laboratory of Computational Intelligence, the director of the Key Laboratory of Cyberspace Big Data Intelligent Security, Ministry of Education. He has been serving as the president of Chongqing Normal University since June 2024. He is the author of more than 10 books, the editor of dozens of proceedings of international and national conferences and has more than 300 reviewed research publications. His research interests include rough sets, granular computing, machine learning, knowledge technology, data mining, neural network, cognitive computing, etc. He was the president of International Rough Set Society (IRSS) 2014–2017, and a council member of the China Computer Federation (CCF) 2008–2023. He is a vice-president with the Chinese Association for Artificial Intelligence (CAAI). He is a fellow of IRSS, CAAI and CCF.
\end{IEEEbiography}

\end{document}